\documentclass[11pt]{article}

\usepackage{acl}

\usepackage{times}
\usepackage{latexsym}

\usepackage[T1]{fontenc}

\usepackage[utf8]{inputenc}

\usepackage{microtype}

\usepackage{booktabs}

\usepackage{amsmath}

\usepackage[title, page]{appendix}

\usepackage{caption}
\usepackage{subcaption}

\usepackage{xfrac}

\usepackage{enumitem}

\usepackage{todonotes}
\setuptodonotes{inline}

\usepackage{microtype}
\usepackage{floatrow}

\usepackage{adjustbox}

\usepackage{placeins}

\usepackage[capitalise]{cleveref}

\usepackage{acro}

\usepackage{multirow}

\PassOptionsToPackage{dvipsnames}{xcolor}
\definecolor{ForestGreen}{RGB}{34,139,34}

\definecolor{plotblue}{RGB}{51, 34, 136}
\definecolor{plotdarkgreen}{RGB}{17, 199, 51}
\definecolor{plotmint}{RGB}{68, 170, 153}
\definecolor{plotlightblue}{RGB}{136, 204, 238}
\definecolor{plotpurple}{RGB}{170, 68, 153}
\definecolor{plotyellow}{RGB}{221, 204, 119}
\definecolor{plotred}{RGB}{204, 102, 119}

\usepackage{tikz}
\usepackage{pgfplots}
\usetikzlibrary{shapes.geometric, arrows, positioning}

\tikzstyle{arrow} = [thick,->,>=stealth]
\pgfplotsset{compat=1.11}

\tikzstyle{start} = [rectangle, 
text width=5cm,
align=center,
text centered,
draw=black]

\tikzstyle{condition} = [rectangle , rounded corners, 
minimum width=2cm, 
minimum height=1cm, 
text centered, 
draw=black, 
fill=gray!10]

\tikzstyle{decision} = [rectangle, 
minimum width=2cm, 
minimum height=1cm, 
text centered, 
draw=black,
yshift= .5cm]

\DeclareAcronym{as}{short = AS , long = acceptance sampling }
\DeclareAcronym{asn}{short = ASN , long = average sample number }

\definecolor{jck}{rgb}{0.8, 0.0, 0.2}

\newcommand{\tikzcircle}[2][red,fill=red]{\tikz[baseline=-0.5ex]\draw[#1,radius=#2] (0,0) circle ;}%

\title{On Efficient and Statistical \\ Quality Estimation for Data Annotation}

\author{Jan-Christoph Klie$^{1}$\thanks{\; Work done while the first author was an intern at Apple.}  \quad Juan Haladjian$^{2}$ \quad Marc Kirchner$^{2}$ \quad Rahul Nair$^{2}$ \\
    $^1$UKP Lab, TU Darmstadt \quad $^2$Apple \\[.2cm]
}

\begin{document}
\maketitle
\begin{abstract}
Annotated datasets are an essential ingredient to train, evaluate, compare and productionalize supervised machine learning models.
It is therefore imperative that annotations are of  high quality.
For their creation, good quality management and thereby reliable quality estimates are needed.
Then, if quality is insufficient during the annotation process, rectifying measures can be taken to improve it.
Quality estimation is often performed by having experts manually label instances as correct or incorrect.
But checking all annotated instances tends to be expensive.
Therefore, in practice, usually only subsets are inspected; sizes are chosen mostly without justification or regard to statistical power and more often than not, are relatively small.
Basing estimates on small sample sizes, however, can lead to imprecise values for the error rate.
Using unnecessarily large sample sizes costs money that could be better spent, for instance on more annotations.
Therefore, we first describe in detail how to use confidence intervals for finding the minimal sample size needed to estimate the annotation error rate.
Then, we propose applying acceptance sampling as an alternative to error rate estimation
We show that acceptance sampling can reduce the required sample sizes up to 50\% while providing the same statistical guarantees.

\end{abstract}

\section{Introduction}

Having large, high-quality annotated datasets available is crucial for successfully training  machine learning models, as well as for evaluating and bringing them into production~\citep{bankoScalingVeryVery2001, sunRevisitingUnreasonableEffectiveness2017, jainOverviewImportanceData2020, sambasivanEveryoneWantsModel2021}.
Therefore, during the creation of new datasets, it is essential to gauge dataset quality efficiently and statistically soundly.
In case the quality is too low, then rectifying measures can be taken to improve it.
These can be, among others, to analyze corner cases, correct annotations, re-train annotators, as well as improving or adjusting the annotation guidelines~\citep{voormannAgileCorpusCreation2008, hovyScienceCorpusAnnotation2010, pustejovskyNaturalLanguageAnnotation2013, ideHandbookLinguisticAnnotation2017}.
It is also important to check dataset quality before release, so that users can rely on the data for their downstream application.

\begin{figure}
    \centering
    \includegraphics[width=0.99\columnwidth]{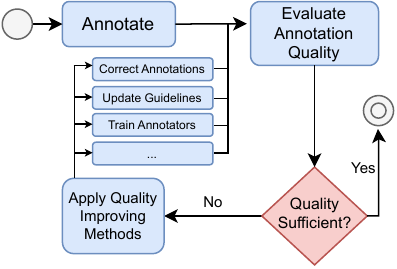}
    \caption{Overview of agile data corpus creation, the recommended workflow to annotate high-quality datasets. This work explores how to efficiently estimate annotation quality using statistics.}
    \label{fig:enter-label}
\end{figure}

\noindent
Estimating annotation quality, however, is expensive, time consuming and tedious, as it typically requires skilled annotators or experts that manually inspect the annotated instances~\citep{monarchHumanintheLoopMachineLearning2021}.
Therefore, in practice, often only small subsamples are inspected and used to estimate data quality or no estimate at all is performed~\citep
{klieAnalyzingDatasetAnnotation2024}.
Small sample sizes  have low statistical power and large error margins, which bears the risk of having inaccurate and too optimistic error rate estimates~\citep{buttonPowerFailureWhy2013, passonneauBenefitsModelAnnotation2014}.
Having no error estimate at all makes it difficult to track data quality throughout the annotation process.
These issues can then result in datasets that contain a non-negligible percentage of incorrect annotations~\citep{northcuttPervasiveLabelErrors2021}.
Subsequently, using datasets with label errors then can lead to inference errors in production, or wrong conclusions when comparing  model architectures and training regimes~\citep{barnesSentimentAnalysisNot2019, reissIdentifyingIncorrectLabels2020, vasudevanWhenDoesDough2022, vadineanuAnalysisImpactAnnotation2022}.

In this work, we hence emphasize the importance of proper quality estimation during, as well as after, the annotation process.
We then describe how to calculate the sample size needed to estimate the proportion of annotation errors  using confidence intervals given certain desired statistical guarantees.
Our analysis shows that estimating the annotation error rate, which is expected to be small, requires relatively large sample sizes to achieve tight bounds and high confidence in the estimate. 
As this can be prohibitively expensive, we suggest using \textit{acceptance sampling}~\citep{dodgeSamplingInspectionPlan1943}, a  statistical quality management technique stemming from industrial manufacturing, instead.
Acceptance sampling does not estimate the error proportion directly.
It judges batches of annotated instances on a accept/reject basis whether they reach a desired quality level or not.
This is done by inspecting subsamples whose size is based on the desired statistical guarantees.
We show that it can be a viable approach for annotation quality estimation: it is able to provide similar statistical guarantees as confidence intervals while requiring up to 50\% less samples.
Our contributions are as follows:

\begin{itemize}[parsep=.25em]
	\item We show the effect of too small sample sizes on the reliability of annotation error estimates. 
	\item We propose using confidence intervals to estimate the annotation error rate efficiently and statistically soundly. 
	\item We propose applying acceptance sampling to the annotation process. Our results show that it is a viable technique for annotation quality control  while reducing required sample sizes compared to confidence intervals.
	\item We provide an easy-to-use and well-tested Python package that implements  confidence intervals and acceptance sampling.\footnote{\url{https://github.com/apple/ml-sampleplan}}
\end{itemize}
\noindent
To the best of our knowledge, we are the first to investigate statistical error rate estimation and acceptance sampling in the context of data annotation.

\section{Background}

Performing annotation quality estimation is crucial during and after the annotation process.
Accurate quality estimates are needed, because if it is known that quality is insufficient, countermeasures can be taken to improve the quality.
This is closely intertwined with how the annotation process is structured, which we discuss first.
Then, we present the most common ways to estimate annotation quality.

\paragraph{Annotation Process}
 
How the annotation process is structured has a large impact on the resulting data quality.
The traditional dataset creation process is similar to the waterfall model, in which phases of data collecting, annotation scheme creation, annotation and validation follow another~\citep{voormannAgileCorpusCreation2008}.
A more modern approach is agile corpus creation~\citep{alexAgileCorpusAnnotation2010, hovyScienceCorpusAnnotation2010, pustejovskyNaturalLanguageAnnotation2013} where annotations are given out in batches.
After each step, the quality is estimated and annotations are validated.
This allows evaluating the quality throughout the annotation process to incorporate feedback, develop or adjust the annotation guidelines, retrain annotators, or re-annotate batches with too many errors~\citep{hovyScienceCorpusAnnotation2010}.
In both cases, it is recommended that annotations are not just published, but the quality is validated via manual inspection during the creation and before release.

\paragraph{Agreement} 

A common approach to determine annotation reliability is to collect multiple annotations per instance. 
These can then be used to compute inter-annotator agreement~\citep{krippendorffReliabilityContentAnalysis2004, artsteinInterCoderAgreementComputational2008}.
However, agreement does not automatically mean correctness, as recent work has shown that state-of-the-art datasets still can contain a non-negligible percentage of errors despite high agreement~\citep{northcuttPervasiveLabelErrors2021}.
Using agreement can also substantially increase the cost.
For many datasets, only one annotation per instance is collected for saving save costs, or because qualified annotators are difficult to recruit.
Then, only a subset of instances are annotated multiple times, based on which agreement is calculated.
Oftentimes, the resulting sample size is too low for a confident estimate with low margin of error~\citep{passonneauBenefitsModelAnnotation2014}.
The computation and interpretation of agreement is often also not straightforward~\citep{amideiAgreementOverratedPlea2019}.
Hence, agreement should not be the only metric of quality~\citep{monarchHumanintheLoopMachineLearning2021}.

\paragraph{Error Rate Estimation} 

Due to the aforementioned issues, it is recommended that annotation projects use (expert) annotators that inspect a (sufficiently large) subset of annotations to estimate the percentage of errors, which we call here the error rate~\citep[Chapter~8.4]{monarchHumanintheLoopMachineLearning2021}.
During agile corpus annotation, the error rate can then be  used to determine whether batches are of sufficient quality or require further improvement.
What is missing in many annotation projects, however, is statistical grounding and analysis of their quality estimation step.
For instance, sample sizes to estimate the error rate are often picked ad-hoc.
That is, they are not grounded in statistics and are often too small~\citep{klieAnalyzingDatasetAnnotation2024}.
The error rate then is often only a point estimate, confidences in the estimates are only rarely taken into account.
This can lead to wrong estimates of the error rate and negatively impact the underlying data quality and downstream usage of the final dataset.

To alleviate the aforementioned issues, we propose using error rate estimation with a sound, statistical footing.
For this, we first define  best practices for estimating the annotation error rate using confidence intervals to determine the minimal sample size needed given desired statistical guarantees.
In addition, we introduce acceptance sampling to the annotation process, which we describe and analyze in the following.
To the best of our knowledge, we are the first to discuss statistical quality control and especially acceptance sampling, coming from manufacturing, to the annotation process.

\paragraph{Automatic Annotation Error Detection}

An alternative to random sampling followed by manual inspection is to use automatic annotation error detection~\citep{dickinsonErrorDetectionCorrection2005}. 
This describes algorithms that automatically find errors, e.g. by using machine learning models or rules. 
Several projects have used annotation error detection to reduce the effort for finding and improving annotated datasets \citep[e.g.,][]{reissIdentifyingIncorrectLabels2020, northcuttPervasiveLabelErrors2021}.
However, it is less suitable for actually estimating annotation quality, because sampling this way might yield biased results and  thus unreliable error estimates.
Therefore, we still find random sampling to be the best tool for precise annotation quality estimation.

\section{Annotation Quality Estimation}

The goal of this work is to ground annotation error estimation in statistics and to make it more economical.
Therefore, in the following, we first model the process of manual inspection as sampling without replacement.
We then describe two statistical quality estimation methods for annotation, \textit{confidence intervals} and \textit{acceptance sampling}.

The setting we consider here is annotation in batches, the modus operandi of the widely recommended, agile annotation process~\citep{voormannAgileCorpusCreation2008, hovyScienceCorpusAnnotation2010, pustejovskyNaturalLanguageAnnotation2013}.
Working with batches instead of all data at once allows stakeholders to incorporate feedback during an annotation project and improve the data quality during annotation if necessary.
But to apply countermeasures in case quality is not sufficient, it is first necessary to know whether the quality is sufficient or not.
As our goal is to estimate annotation quality with a minimum of manual inspection for a given level of confidence and precision, we also discuss how to compute the sample size needed for each method.
We compare the presented methods in \cref{sec:results}.

\subsection{Quality Estimation as Sampling}

We model the process of manually inspecting a subset of annotations as sampling $n$ instances from a dataset of size $N$ and labeling them as either correct or incorrect.
Because no instance is inspected repeatedly, we model it as sampling without replacement.
Hence, this is best described by the hypergeometric distribution.
Note that our sampling is independent of the initial annotators and their chance of making errors, as the inspection happens after the annotations have been already made.

Approximating the hypergeometric distribution with a binomial distribution is tempting, as it would simplify the math and its implementation.
But in our setting, this is often not justified, as the rule of thumb for it, $\sfrac{n}{N} < 0.1$, does not hold; the batch size is relatively small and for tight statistical guarantees, the sample size is comparatively large.
Using this approximation can lead to wildly different and often higher suggested sample sizes compared to the exact solution.
This is also shown in supplementary results (\cref{app:binomial}).
Using the hypergeometric distribution, however, also makes the following implementations more complicated and error-prone, which is why we employ testing and validation for our accompanying Python library.
But one advantage when using the hypergeometric distribution is that it has a natural, maximum number of inspections, which is when each instance has been inspected once.
This is not so clearly defined for the binomial setting as items can be inspected more than once.

Our model assumes that instances and errors are made independent and are identically distributed.
Therefore, it is important that the actual sampling procedure and granularity takes this into account.
For instance for span labeling, (an example task would be named entity recognition), the sampling should be based at least on sentences, as spans often depend on each other.
For text recognition or object classification datasets, an instance should preferably be an image and not a single bounding box.

\subsection{Error Rate Estimation}
\label{sec:method_error_rate}

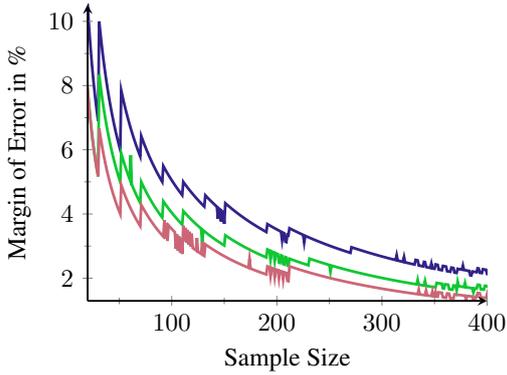
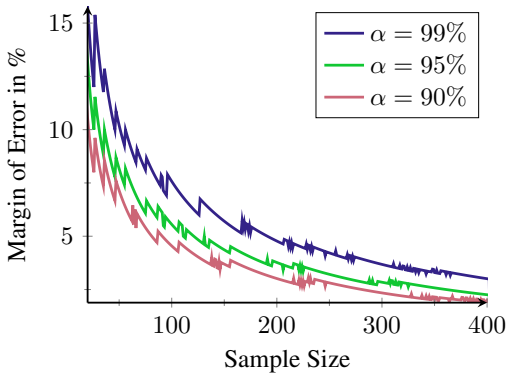
\begin{figure}[t]
    \centering
    \captionsetup[subfigure]{justification=centering}

    \begin{subfigure}[b]{0.99\textwidth}
        \centering
        \resizebox{.9\textwidth}{!}{
\begin{tikzpicture}
\begin{axis}[
    axis lines = left,
  axis line style = thick,
      width=\textwidth,
    height=0.8\textwidth,
  xmin = 20,
  xmax = 400,
  axis on top,
  minor x tick num=1,
  minor y tick num=1,
  xlabel = Sample Size,
  ylabel = Margin of Error in \%,
]

\addplot [color=plotblue, very thick] table [x=x, y=99, col sep=comma] {data/sample_size_precision_5.csv};

\addplot [color=plotdarkgreen,  very thick] table [x=x, y=95, col sep=comma] {data/sample_size_precision_5.csv};

\addplot [color=plotred,  very thick] table [x=x, y=90, col sep=comma] {data/sample_size_precision_5.csv};

\end{axis}

\end{tikzpicture}}
        \caption{$p_e = 0.05$}
        \label{fig:sampling_error_5}
    \end{subfigure}
    \\
    \begin{subfigure}[b]{0.99\textwidth}
        \centering
        \resizebox{.9\textwidth}{!}{
\begin{tikzpicture}
\begin{axis}[
    axis lines = left,
  axis line style = thick,
      width=\textwidth,
    height=0.8\textwidth,
  xmin = 20,
  xmax = 400,
  axis on top,
  minor x tick num=1,
  minor y tick num=1,
  xlabel = Sample Size,
  ylabel = Margin of Error in \%,
]

\addplot [color=plotblue, very thick] table [x=x, y=99, col sep=comma] {data/sample_size_precision_10.csv};
\addlegendentry{$\alpha=99\%$}

\addplot [color=plotdarkgreen,  very thick] table [x=x, y=95, col sep=comma] {data/sample_size_precision_10.csv};
\addlegendentry{$\alpha=95\%$}

\addplot [color=plotred,  very thick] table [x=x, y=90, col sep=comma] {data/sample_size_precision_10.csv};
\addlegendentry{$\alpha=90\%$}

\end{axis}
\end{tikzpicture}}
        \caption{$p_e = 0.1$}
        \label{fig:sampling_error_10}
     \end{subfigure}

    \caption{Sampling error vs. margin of error when sampling without replacement for manually inspecting a dataset to estimate the error rate. We compute a hypergeometric confidence interval for different confidence levels $\alpha$ and two underlying, true error rates $p_e$ and $N=1000$. The closer the true error rate (and thereby hopefully the assumed error rate to compute the sample size)  is to $0.5$, the larger the required sample size is. The jaggedness is caused by the distribution's discreteness.}
    \label{fig:sampling_error}
\end{figure}

The most commonly used approach to quality estimation is to directly estimate the error rate.
It is computed as the ratio of annotations having incorrect labels and the total number of annotations.
If it is too high for a batch of annotated instances, then it is rejected and measures are to be taken to improve its quality. 
In the vast majority of works that actually compute the error rate, it is given as a point estimate~\citep{klieAnalyzingDatasetAnnotation2024}. 
This has the disadvantage that it does not convey a notion of precision or potential margin of error (see also \cref{fig:sampling_error}).
Therefore, we suggest using confidence intervals, which are a common method to estimate a proportion and gauge the certainty of the estimate.

A confidence interval is an interval estimate for an unknown parameter $\theta$.
Its width is given by a designated confidence level $\alpha$.
In the long run, when estimating $\theta$ repeatedly from numerous samples, then approximately $\alpha$\% of the computed confidence intervals contain $\theta$.
Desirable properties of confidence intervals are that the specified $\alpha$-level actually holds while being as narrow as possible; the width of the confidence interval determines the precision of an estimate.
An interval's margin of error (also called half-width, it is equal to half of an interval's width) is a more intuitive value to express precision: given a point estimate $\hat{\theta}$ and a margin of error $m$, if sampling repeatedly many times, then $\alpha$\% of estimates will be in $[\hat{\theta} - m;\ \hat{\theta} + m]$.

We favor using confidence intervals over point estimates, as the former give additional information about confidence and precision.
Confidence intervals can also be used to estimate the sample size needed for a given confidence level, which we investigate in our analysis.
We are not aware of any works using confidence intervals to estimate the error rate or required sample size.
While checking the literature, we encountered only few works that reported an error rate.
All of which used point estimates and did not justify the sample size in case they inspected only a subset.

Many different ways of constructing confidence intervals have been devised for estimating proportions~\citep{brownIntervalEstimationBinomial2001}.
Using confidence intervals based on uncorrected normal approximations, like the Wald interval that is often taught in many statistics textbooks and courses, is strongly discouraged as it has low coverage and performs poorly for proportions close to $0$ and $1$~\citep{santnerNoteTeachingBinomial1998, brownIntervalEstimationBinomial2001}.
But low error rates are desired and expected for annotation errors.
Also, most confidence intervals are designed for the binomial distributions and can again give incorrect answers when using them for hypergeometric distributions.
This is why we recommend computing an the exact confidence interval that directly uses the underlying distribution \citep{clopperUseConfidenceFiducial1934}.
An exact $(1 - \alpha)$ confidence interval for a parameter $\theta$ modeled by a discrete distribution, as proposed by \citet{clopperUseConfidenceFiducial1934}, is given by
\begin{align*}
    P(X \geq k; \theta_l) &= \frac{\alpha_1}{2} \ , \\
    P(X \leq k; \theta_u ) &= \frac{\alpha_2}{2}\ , \\
    \alpha_1 + \alpha_2    &\leq \alpha \ ,
\end{align*}
where $P(\cdot)$ is a cumulative distribution function. 
The result after solving this equation system is an confidence interval $[\theta_l, \theta_u]$ where $\theta_l \leq \theta \leq \theta_u$.
For simplicity, similarly as in the literature, we assume that $\alpha_1 = \alpha_2$.
Then, the equations can be solved individually to find $\theta_l$ and $\theta_u$ by numerically finding roots in $P_l(\theta_l) - \sfrac{\alpha}{2} = 0$ and $P_u(\theta_u) - \sfrac{\alpha}{2} = 0$.

To compute the sample size required for a given interval half-width and confidence level $\alpha$, we similarly solve
\begin{equation*}
    P(X \geq k; \theta_l) + P(X \leq k; \theta_u ) - \alpha = 0 \ ,
\end{equation*}

\noindent
via numerical root finding.
This requires assuming a parameter $\hat{\theta}$ (in this work, $\theta$ is an assumption for the error rate)  as well as the desired interval width, from which $\theta_l$ and $\theta_u$ are computed.

Due to the underlying discreteness of the hypergeometric distribution, this system of equations is usually not exactly solvable for every desired value of $\alpha$.
As a side effect, exact intervals are conservative, that means they have higher confidence, coverage and width than specified for a $(1 - \alpha)$ interval~\cite{newcombeTwosidedConfidenceIntervals1998, brownIntervalEstimationBinomial2001}.
Even though this incurs additional costs, we argue that it is desirable that an exact interval errs on the side of suggesting slightly larger sample sizes than required and working for any proportion.
Approximations would instead provide too small sample sizes or might break down in edge cases.
For the hypergeometric distribution, more complex algorithms exist that can find minimal-width intervals~\citep{wangExactOptimalConfidence2015, bartroffOptimalFastConfidence2022}.
These are difficult to implement, very few software packages for them exist as of yet and their inversion to compute the sample size for given confidence levels and width instead of the confidence interval itself is not given.
This is why we leave their application as future work but point them out as potential alternatives to using exact Clopper-Pearson intervals.

\subsection{Acceptance Sampling}

\begin{figure}[t]
    \centering
     \captionsetup[subfigure]{aboveskip=-1pt,belowskip=-5pt}

    \begin{minipage}{.48\linewidth}
        \begin{subfigure}[t]{.99\linewidth}
            \caption{Single Sampling}
            \resizebox{\textwidth}{!}{
\begin{tikzpicture}[node distance=2cm]
\node (start) [start] {Inspect a random sample of size $n$ and count the number of incorrect instances $d$};
\node (shouldaccept) [condition, below left of=start] {$d \le c$};
\node (shouldreject) [condition, below right of=start] {$d > c$};
\node (accept) [decision, below of=shouldaccept,fill=green!30, yshift=.2cm] {Accept};
\node (reject) [decision, below of=shouldreject,fill=red!30, yshift=.2cm] {Reject};

\draw [arrow] (start) -- (shouldaccept);
\draw [arrow] (start) -- (shouldreject);

\draw [arrow] (shouldaccept) -- (accept);
\draw [arrow] (shouldreject) -- (reject);
\end{tikzpicture}}
            \label{fig:as_single_sampling}
        \end{subfigure} \\
        \begin{subfigure}[t]{.99\linewidth}
            \caption{Sequential Sampling}
            \resizebox{\textwidth}{!}{
\begin{tikzpicture}
  \begin{axis}[
  axis line style = thick,
  enlargelimits=true,  
  axis on top,
  ymin=0,ymax=5.5,
  xmin=0,xmax=10,
  axis equal=true,
  unit rescale keep size=false,
  xlabel=Cumulative Sample Size,
  ylabel=\# of incorrect instances,
  xticklabel=\empty,
  yticklabel=\empty,
  minor tick num=1,
  axis lines = middle,
  x label style={at={(axis description cs:0.5,-0.03)},anchor=north, font=\small},
  y label style={at={(axis description cs:-0.03,.5)},rotate=90,anchor=south, font=\small},
  ]
    \addplot[fill=red!30, color=red!30] coordinates {(0, 2) (10, 6) (0, 6) (0,2)};
    \addplot[fill=green!30, color=green!30] coordinates {(2.5, 0) (10, 3) (10, 0) (2.5,0)};
    \addplot[fill=gray!10, color=gray!10] coordinates {(0,0) (2.5, 0) (10, 3) (10, 6) (0,2) (0, 0)};
    \addplot[color=blue, thick] coordinates {(0,0) (1,0) (1,0) (1,1) (2,1)  (2,2) (3,2) (4,2) (4,3) (5,3) (5,4) (6,4) (7,4) (7,5)   };

        \addplot [thick,color=red,mark=o,fill=red, 
                    fill opacity=0.05]coordinates {
            (7, 5)  };

    \node[] at (axis cs: 2.5,4.5) {Reject};
    \node[] at (axis cs: 8,1) {Accept};
    \node[] at (axis cs: 6,2.5) {Continue};

  \end{axis}
\end{tikzpicture}}
            \label{fig:as_sequential}
        \end{subfigure} 
    \end{minipage}
    \hfill
    \begin{minipage}{.48\linewidth}
        \begin{subfigure}[t]{.99\linewidth}
            \vspace{-1em}
            \caption{Double Sampling}
            \resizebox{1\textwidth}{!}{
\begin{tikzpicture}[node distance=2cm]

\node (start) [start] {Inspect a random sample of size $n_1$ and count the number of incorrect instances $d_1$};

\node (shouldaccept1) [condition, below left of=start] {$d_1 \le c_1$};
\node (shouldcontinue) [condition, below of=start, yshift=-.6cm] {$c_1 < d_1 \le c_2$};
\node (shouldreject1) [condition,  below right of=start] {$d_1 > c_2$};

\node (accept) [decision, below left of=shouldcontinue,fill=green!30, yshift=-.25cm] {Accept};
\node (reject) [decision, below right  of=shouldcontinue,fill=red!30, yshift=-.25cm] {Reject};

\node (shouldaccept2) [condition, below of=accept, yshift=.75cm] {$d_1 + d_2 \le c_1$};
\node (shouldreject2) [condition, below of=reject, yshift=.75cm] {$d_1 + d_2 > c_2$};

\node (secondsample) [start, below right of=shouldaccept2, yshift=-.3cm] {Inspect a second random sample of size $n_2$ and count the number of incorrect instances $d_2$};

\draw [arrow] (start) -- (shouldaccept1);
\draw [arrow] (start) -- (shouldreject1);

\draw [arrow] (shouldaccept1) -- (accept);
\draw [arrow] (shouldreject1) -- (reject);

\draw [arrow] (shouldaccept2) -- (accept);
\draw [arrow] (shouldreject2) -- (reject);

\draw [arrow] (shouldaccept2) -- (accept);
\draw [arrow] (shouldreject2) -- (reject);

\draw [arrow] (start) -- (shouldcontinue);
\draw [arrow] (shouldcontinue) -- (secondsample);

\draw [arrow] (secondsample) -- (shouldaccept2);
\draw [arrow] (secondsample) -- (shouldreject2);
\end{tikzpicture}}
            \label{fig:as_double_sampling}
        \end{subfigure}
    \end{minipage}
    \vspace{-2em}
    \caption{Flowcharts for the three different acceptance sampling methods discussed in this work.}
    \label{fig:acceptance_sampling}    
\end{figure}
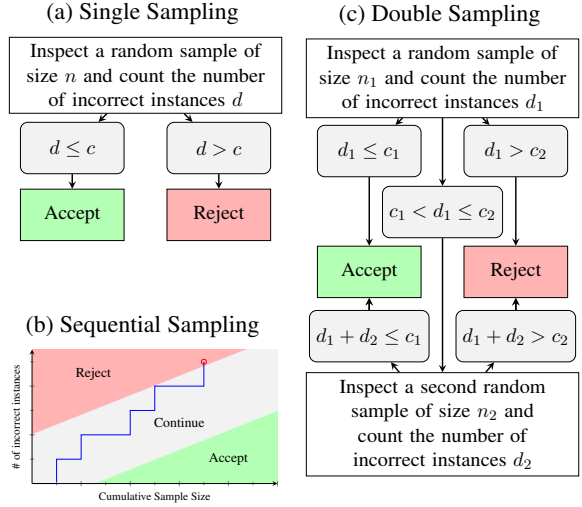

Acceptance Sampling~\citep{dodgeSamplingInspectionPlan1943} is a quality management approach originally applied by the US military to gauge the quality procured equipment on delivery.
Adjusted for the task of estimating the quality of an annotated dataset, it works the following.
Each item in a batch is either a correctly or incorrectly annotated instance.
Instead of inspecting 100\% of each batch, a subsample of instances is taken and manually inspected.
If too many instances are found to be incorrect, then the whole batch is rejected, otherwise it is accepted.

A \textit{sampling plan} defines how many instances to sample, how to sample, and the boundaries for when to accept or reject.
It is the solution to the following equations:
\begin{align}
    \label{eq:1}
    P(X \leq c; p_a, n) &\geq 1 - \alpha \ , \\
    P(X \leq c; p_r, n) &\leq \beta \ , \nonumber
\end{align}
with
\begin{equation*}
    0 < p_a < p_r < 1 \ ,
\end{equation*}
and
\begin{equation*}
    1 > 1- \alpha > \beta > 0 .
\end{equation*}
The probability $P$ is dependent on the underlying distribution and sampling procedure.
In general, \cref{eq:1} is non-linear and has no closed-form solution.
Inputs are the predetermined desirable acceptance error rate $p_a$ and unacceptable error rate $p_r$ as well as the so called producer's risk $\alpha$ and consumer's risk $\beta$.
$\alpha$ is the probability to reject a good batch of instances, $\beta$ the probability to accept a bad batch.
These are similar to Type I (false-positive) and Type II errors (false-negative) and $\alpha$ is identical to the $\alpha$ for confidence intervals.
The range between $p_a$ and $p_r$ is called the \textit{indifference zone}, we assume that quality in this zone is good enough for  borderline accepting a batch.
Each plan has an associated average sample number, that is, on average, how many instances need to be inspected until a verdict to accept or reject is reached.

It has to be noted that the goal of acceptance sampling is not to estimate the actual error rate, it just determines when to accept or reject a batch.
But if the batch is accepted, then its latent error rate is below $p_a$ with a chance of $\beta$.
In the following, we discuss the three most common acceptance sampling variants, single sampling, double sampling and sequential sampling.
A depiction of the different acceptance sampling types is given in \cref{fig:acceptance_sampling}.

\paragraph{Single Sampling} 

From a batch of annotated instances of size $N$, a sample of size $n$ is inspected.
If the sample contains more annotation errors than a critical value $c$, the whole batch is rejected, otherwise, it is accepted.
\noindent
For single sampling, \cref{eq:1} can be solved via systematic search, as described by \citet{guentherUseBinomialHypergeometric1969} and \citet{meekerSequentialTestsHypergeometric1975}.

\begin{enumerate}[itemsep=.25em]
    \item Start with $c^* = 0$.
    \item Find the largest $n_U$ so that \\$P(X \leq c^*; p_a, n_U) \geq 1 - \alpha$
    \item Find the smallest $n_L$ so that \\$P(X \leq c^*; p_r, n_L) \leq \beta$
    \item If $n_L \leq n_U$, then $(n_L, c^*$ is a plan that satisfies the requirements with minimum sample size, $n^* = n_L$. 
    \item If $n_L > n_U$, increment $c^*$ by one and go to step 2.
\end{enumerate}

\noindent
Step 2 and 3 can be computed via the quantile function of the respective distribution.
The average sample number for a plan is $n^*$ and constant.

\paragraph{Double Sampling} Instead of taking only a single sample, in double sampling, batches are accepted or rejected based on two consecutive, smaller subsamples.
At first, a sample of size $n_1$ is taken and inspected.
If it contains less incorrect annotations than a lower limit $c_1$, the batch is accepted, if it contains more error than an upper limit $c_2$, the batch is rejected.
If the number of errors it is between both, then a second sample of size $n_2$ is taken; the batch is rejected if the number of incorrect instances in both samples combined is larger than $c_2$. 
The advantage of double sampling is that in the happy case, only $n_1$ samples need to be inspected, thereby saving time and money. 
To make the actual computation more tractable, we use only double-stage plans where $n_1 = n_2$ . 
We analyze two versions of double sampling, \textit{full} where samples are always completely inspected and \textit{curtailed}, where inspection of the second sample is stopped in case there are more than $c_2$ defects found.
It is recommended to always at least look at the first $n_1$ samples to get a rough estimate for the error rate~\citep{duncanQualityControlIndustrial1959}, we will follow this textbook advice in this work.

Double sampling plans as solutions to \cref{eq:1} can be computed via Algorithm 1 in \citet{lucaWebbasedToolDesign2020}, to which we refer the interested reader.
The average sample number for a full double sampling plan is then given by
\begin{equation*}
    \mbox{ASN}_{f} = n_1 \cdot P_I + (n_1 + n_2) (1 - P_I) \ ,
\end{equation*}
\noindent
where $P_I$ is the probability that the batch is accepted or rejected after the first sample, 
\begin{equation*}
    P_I = P(D_1 \leq c_1) + P(D_1 > c_2) .
\end{equation*}
\noindent
The ASN for a curtailed double sampling plan is 
\begin{align*}
    \mbox{ASN}_{c} = n_1 & + \sum_{i=c_1 + 1}^{c2} P(n_1, i) [ n_2  P_L(n_2, c_2 - i) \\ 
    &+ \frac{c_2 - i + 1}{p} P_M(n + 1, c_2 - i + 2) ]  \ ,
\end{align*}
\noindent
where $P(n_1, i)$ is the probability of observing exactly $i$ errors in a sample of size $n_1$, $ P_L(n_2, c_2 - i)$ the probability of observing $c_2 - i$ or fewer errors in in a sample size of $n_2$ and $P_M(n + 1, c_2 - i + 2)$ the probability of observing exactly $c_2 - i + 2$ errors or more in a sample of size $n + 1, c_2$.
A good description of double sampling and its mathematical description can also be found in \citet[][Part 2 VIII 1.3]{duncanQualityControlIndustrial1959} and \citet[][Section 15.3.1]{montgomeryIntroductionStatisticalQuality2013}.

\paragraph{Sequential Sampling} 

The generalization of double sampling is sequential sampling.
It is based on the sequential probability ratio test~\citep{waldSequentialAnalysis1947, meekerSequentialTestsHypergeometric1975}.
In this setting, instances in a batch are inspected one by one and after each step, it is decided whether to stop and accept or reject a batch, or to continue the inspection.
The acceptance and rejection boundaries  are computed at every step from $p_a$ and $p_r$, $\alpha$ and $\beta$ as well as the number of incorrect and total instances inspected so far.
It can happen that the stopping criterion never would hold and thereby the whole batch would need to be inspected, especially if the actual error rate is between $p_a$ and $p_r$.
As this is an undesirable outcome, we recommend stopping inspection at the sample size of single sampling and accept or reject based on its critical value, as proposed by ~\citet{meekerSequentialTestsHypergeometric1975}.
Thereby, the statistical guarantees still hold, as then sequential sampling essentially turned into single sampling.
Curtailment requires adjusting the acceptance and rejection borders to shrink the zone of indifference and thus stop earlier.
Note that this curtailment and adjustment is an approximation and better stopping conditions can exist.
But computing an optimal curtailment in general is difficult~\citep{tantaratanaTruncatedSequentialProbability1977}, which is why we leave evaluating this case for future work.

For the binomial case, closed form solutions for \cref{eq:1} exist to compute sequential sampling plans, see \citet[][Part 2 VIII 2.1]{duncanQualityControlIndustrial1959} or \citet[][15.3.3]{montgomeryIntroductionStatisticalQuality2013}.
For the hypergeometric distribution, computing a sequential sampling plan is more intricate.
We use the direct method proposed by \citet{aroianSequentialAnalysisDirect1968} and adapted by \citet{meekerSequentialTestsHypergeometric1975} for the hypergeometric distribution.
When curtailing, for hypergeometric distribution, we compute the threshold via single sampling and  wedge truncation, for the binomial, we use $3n$ from the corresponding single sampling plan~\citep{montgomeryIntroductionStatisticalQuality2013}.

\section{Sample Size Analysis}
\label{sec:results}

To make quality management more feasible, our goal is to estimate the quality in annotated batches given a desired set of statistical guarantees while inspecting as few samples as possible.
To emphasize the importance of choosing a proper sample size, we first briefly simulate the relation between sample size and the reliability of an estimate.
If it is too small, then the margin of error is large compared to the proportion estimated and the quality estimate cannot be trusted.
Then, we compare confidence intervals and acceptance sampling  with regard to their required \ac{asn}.

The following results are derived analytically and are agnostic to the underlying datasets, as long as the statistical assumptions hold; only dataset size and error counts are used.
We show the application of acceptance sampling on existing datasets with known error rate~\cref{sec:app_real}.

The \ac{asn} for a method is the expected number of instances that need to be inspected until a decision can be made to accept or reject a batch.
In case of confidence intervals and single sampling, the \ac{asn} is a constant; the whole sample is always inspected as there is no form of early stopping.
For double and sequential sampling, the \ac{asn} depends on the true error rate in the current batch.

\begin{table}[t]

\centering
\begin{tabular}{@{}crr@{}}
\toprule
\textbf{}     & \textbf{Strict} & \textbf{Relaxed} \\ \midrule
$p_a$          & 0.01            & 0.02             \\
$p_r$          & 0.03            & 0.05             \\
$\alpha$         & 0.01            & 0.05             \\
$\beta$          & 0.1             & 0.2              \\
CI half width & 0.01            & 0.02             \\ \bottomrule
\end{tabular}
\label{tab:qa_configs}
\caption{Target quality levels}

\end{table}

For our analysis, we assume a batch size of $N = 1000$ .
If it was smaller, then it is often feasible to inspect whole batches.
If it was larger, then this causes only minor changes to the required sample sizes (see \cref{fig:asn_p_binom}).
We choose two sets of values simulating two different quality control regimes, which we call \textit{strict} and \textit{relaxed}. 
They are listed \cref{tab:qa_configs}.
Computing the minimum sample size to achieve a certain confidence interval width given a confidence level requires assuming a value for the a-priori unknown true error rate $p_e$. 
We choose to calculate the sample size both for $p_e = p_a$ as well as $p_e = p_r$.
The closer $p_e$ is to $0.5$, the larger the required sample size would be.
Note that using a sample size estimate for a $p_e$ leads to underpowered estimates if $p > p_e$, which makes using confidence intervals more difficult to use correctly compared to acceptance sampling, where this is not needed.
Therefore, the rates for which we estimate confidence interval sample sizes can be seen as an optimistic estimate.

\subsection{Sampling Error}

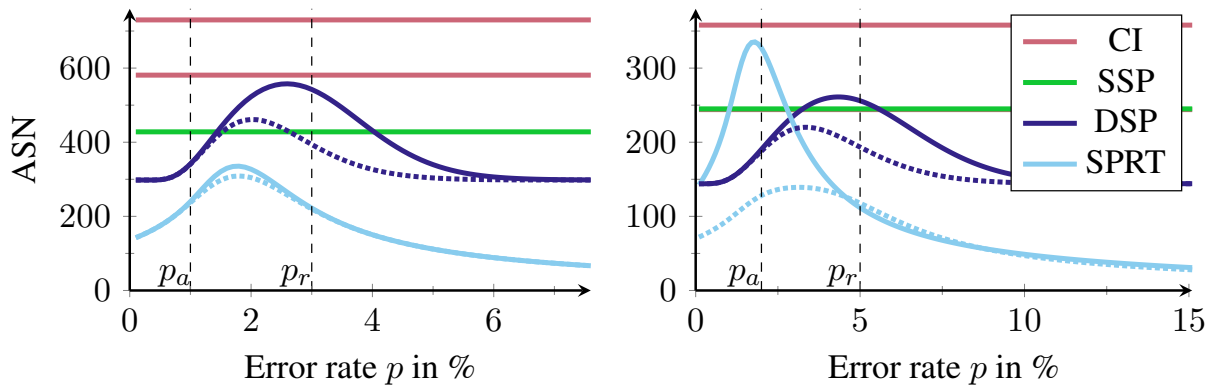
\begin{figure*}[t]
    \centering
    \captionsetup[subfigure]{justification=centering}

    \begin{subfigure}[t]{0.495\textwidth}
        \centering
        \resizebox{!}{.65\textwidth}{
\begin{tikzpicture}
\begin{axis}[
    axis lines = left,
  axis line style = thick,
    ylabel = ASN,
    axis on top,
    xlabel = Error rate $p$ in \%,
    minor x tick num=1,
    minor y tick num=1,
    xmin = 0,
    width=0.85\textwidth,
    height=0.6\textwidth,
]

\addplot [color=plotred, ultra thick] table [x=p, y=CI hypergeometric p1, col sep=comma] {data/raw_asns_strict.csv};

\addplot [color=plotdarkgreen, ultra thick] table [x=p, y=SSP hypergeometric, col sep=comma] {data/raw_asns_strict.csv};

\addplot [color=plotblue, ultra thick] table [x=p, y=DSP hypergeometric full, col sep=comma] {data/raw_asns_strict.csv};

\addplot [color=plotlightblue, ultra thick] table [x=p, y=SPRT hypergeometric full, col sep=comma] {data/raw_asns_strict.csv};

\addplot [color=plotred, ultra thick] table [x=p, y=CI hypergeometric p2, col sep=comma] {data/raw_asns_strict.csv};

\addplot [color=plotblue, densely dotted, ultra thick] table [x=p, y=DSP hypergeometric curtailed, col sep=comma] {data/raw_asns_strict.csv};

\addplot [color=plotlightblue, ultra thick, densely dotted] table [x=p, y=SPRT hypergeometric curtailed, col sep=comma] {data/raw_asns_strict.csv};

\addplot +[mark=none, color=black, dashed] coordinates {(1, 0) (1, 760)};
\addplot +[mark=none, color=black, dashed] coordinates {(3, 0) (3, 760)};

\node[] at (axis cs: 0.75,40) {$p_a$};
\node[] at (axis cs: 2.75,40) {$p_r$};

\end{axis}
\end{tikzpicture}}
        \caption{Strict configuration. The SSP has $n =428, c = 8$, the DSP $n_1=n_2=298, c_1=4, c_2 =11$.}
        \label{fig:asn_p_strict}
     \end{subfigure}
     \hfill
    \begin{subfigure}[t]{0.495\textwidth}
        \centering
        \resizebox{!}{.65\textwidth}{
\begin{tikzpicture}
\begin{axis}[
    axis lines = left,
  axis line style = thick,
  xmin = 0,
  axis on top,
  minor x tick num=1,
  minor y tick num=1,
  width=0.9\textwidth,
  height=0.6\textwidth,
  xlabel = Error rate $p$ in \%,
]

\addplot [color=plotred, ultra thick] table [x=p, y=CI hypergeometric p1, col sep=comma] {data/raw_asns_relaxed.csv};
\addlegendentry{CI}

\addplot [color=plotdarkgreen, ultra thick] table [x=p, y=SSP hypergeometric, col sep=comma] {data/raw_asns_relaxed.csv};
\addlegendentry{SSP}

\addplot [color=plotblue, ultra thick] table [x=p, y=DSP hypergeometric full, col sep=comma] {data/raw_asns_relaxed.csv};
\addlegendentry{DSP}

\addplot [color=plotlightblue, ultra thick] table [x=p, y=SPRT hypergeometric full, col sep=comma] {data/raw_asns_relaxed.csv};
\addlegendentry{SPRT}

\addplot [color=plotred, ultra thick] table [x=p, y=CI hypergeometric p2, col sep=comma] {data/raw_asns_relaxed.csv};

\addplot [color=plotblue, densely dotted, ultra thick] table [x=p, y=DSP hypergeometric curtailed, col sep=comma] {data/raw_asns_relaxed.csv};

\addplot [color=plotlightblue, ultra thick, densely dotted] table [x=p, y=SPRT hypergeometric curtailed, col sep=comma] {data/raw_asns_relaxed.csv};

\addplot +[mark=none, color=black, dashed] coordinates {(2, 0) (2, 380)};
\addplot +[mark=none, color=black, dashed] coordinates {(5, 0) (5, 380)};

\node[] at (axis cs: 1.5,20) {$p_a$};
\node[] at (axis cs: 4.5,20) {$p_r$};

\end{axis}
\end{tikzpicture}}
        \caption{Relaxed configuration. The SSP has $n =425, c = 8$, the DSP $n_1=n_2=144, c_1=3, c_2 =9$.}
        \label{fig:asn_p_relaxed}
     \end{subfigure}
    \caption{Average sample numbers (ASN) required for a strict and relaxed configuration for \textbf{C}onfidence \textbf{I}ntervals (CI), \textbf{S}ingle \textbf{S}ampling \textbf{P}lans (SSP), \textbf{D}ouble \textbf{S}ampling \textbf{P}lans (DSP), and Sequential Sampling Plans  based on the \textbf{S}equential \textbf{P}robability \textbf{R}atio \textbf{T}est (SPRT). Dotted lines are plans with curtailment. The confidence interval requiring the smaller sample size is the one assuming $p_a$.}
    \label{fig:asn_p}
\end{figure*}

In \cref{{fig:sampling_error}}, we show the relation between sample sizes and the precision of the resulting estimate.
The smaller a sample that is used to estimate a proportion, here the error rate, the more imprecise an estimate is.
As a proxy for the reliability of the estimate, we compute a hypergeometric confidence interval and use the resulting margin of error as our metric, which is half of the interval's width (\cref{sec:method_error_rate}).
We assume here an underlying true error rate in this simulation of $5\%$ or $10\%$, which are common error values based on the literature~\citep{northcuttPervasiveLabelErrors2021}.
Overall, it can be seen that for sample sizes up to around $300$, the margin of error is relatively large. 
An interval estimate for instance based on 100 samples would be $5\pm 4\%$ for the former and $10\pm 5.5\%$ for the latter with 95\% confidence.
In case the sample size is 200, then it would be $5\pm 2.75\%$ for the former and $10\pm 3.75\%$ for the latter with 95\% confidence.
This shows that for lower sample sizes, the error margin causes the estimates to be rather imprecise and not useful.

\subsection{Average Sample Sizes}

In the following, we analyse how many samples are needed at minimum for given statistical guarantees and desired quality levels.
The \ac{asn} curves for the strict and relaxed configurations are shown in \cref{fig:asn_p}.
Note that these are obtained analytically and are not simulated.
It can be seen that actually estimating the error rate with confidence intervals  almost always requires the largest sample sizes and around double the samples compared to the best acceptance sampling approach, curtailed sequential sampling.
This is because the error rate is assumed to be a relatively small proportion and a small interval width is desired in order to have meaningful bounds.
Double sampling without curtailment typically requires the same or more samples than single sampling, but curtailment almost always has the same or better sample efficiency.
Sequential sampling requires overall the fewest samples when used with curtailment.
Settling with less confidence in the results, that is using the \textit{relaxed} configuration, can half the sample size needed compared to a stricter plan. 

Overall, we see that acceptance sampling, while not directly estimating the error rate, can be a feasible and economic alternative compared to the typically used point estimate or confidence intervals.
It can reduce the required sample size by almost half while providing the same statistical guarantees.

\section{Acceptance Sampling in Practice}
\label{sec:app_real}

In the following, we show the application of acceptance sampling on actual datasets with known error rates and observe their behaviour.
For this, we use two datasets that have been found to contain non-negligible amounts of annotation errors:

\begin{description}[itemsep=-1ex]
\item[CoNLL 2003] is a dataset for named entity recongition, introduced by \citet{tjongkimsangIntroductionCoNLL2003Shared2003}; errors analysis was conducted by \citet{reissIdentifyingIncorrectLabels2020}.
\item[IMDB] is a dataset for sentence-level sentiment classification, introduced by \citet{maas-etal-2011-learning}, error analysis was performed by \citet{northcuttPervasiveLabelErrors2021}.
\end{description}

\begin{figure}[t]
    \centering
    \includegraphics[width=0.99\textwidth]{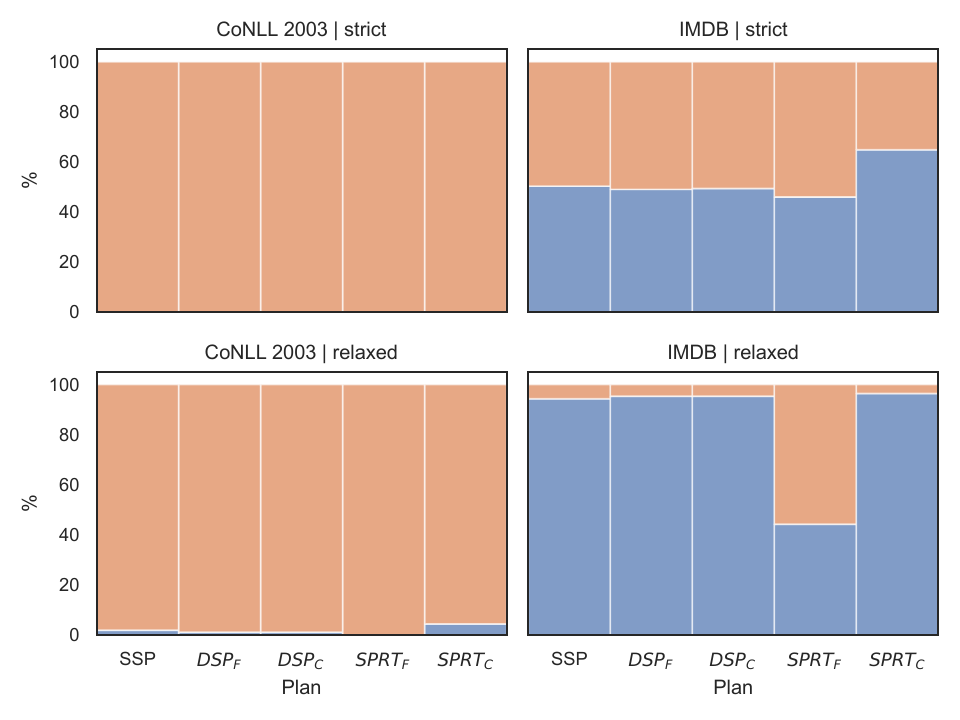}
    \caption{Simulating using acceptance sampling on existing NLP datasets. We run 1000 simulations with different seeds and count how often a sample was accepted \tikzcircle[fill=blue]{3pt} or rejected \tikzcircle[fill=orange]{3pt}.}
    \label{fig:real-world}
\end{figure}

\noindent
We use the dataset and error counts as described in \citet{klieAnnotationErrorDetection2023}.
These are listed in \cref{tab:practice_stats}.

\begin{table}[b]
    \begin{center}
    \resizebox{.99\textwidth}{!}{
    \begin{tabular}{@{}lccc@{}}
    \toprule
    \textbf{Dataset} & \textbf{\#Errors} & \textbf{\#Sentences} & \textbf{\% Error} \\ \midrule
    ConLL 2003       & 217               & 3380                 & 6.42                   \\
    IMDB             & 499               & 24799                & 2.01                   \\ \bottomrule
    \end{tabular}}
    \end{center}
    \caption{Dataset and error statistics for ConLL 2003 and IMDB as given in \citet{klieAnnotationErrorDetection2023}.}
    \label{tab:practice_stats}
\end{table}

\noindent
For CoNLL 2003, we consider annotations and errors on sentence-level in order to satisfy the independence assumption needed for the statistical methods used; a sentence is considered erroneous if at least one span is annotated incorrectly.
We use the same configurations as described in \cref{tab:qa_configs}.

\noindent
\cref{fig:real-world} shows the results. 
It can be seen that for Conll-2003, almost all samples are rejected. This is expected, because the error rate is above 3\% and 5\% for the strict and relaxed configurations, respectively.
For IMDB, the error rate is in the zone of indifference for the strict configuration. Thus, the results are inconclusive.
In a real setting, these batches would be either accepted or inspected further to determine the underlying error rate as well as to correct instances~\citep[][Chapter 6]{NAP372}.
In its relaxed configuration, most samples are accepted even though the error rate is slightly above $p_a$.
Most methods still accept there, except for sequential sampling without curtailment.
Average sample sizes for these plans are listed in \cref{tab:real-world}.

\section{Conclusion}

In this work, we proposed two methods statistical concerning quality control for data annotation.
First, we showed how to estimate the error rate statistically sound using confidence intervals.
Given desired confidence and  precision of the estimate, confidence intervals compute the minimal sample sizes needed to statistically guarantee these.
Hence, we presented an alternative technique from manufacturing, that is, acceptance sampling.
Acceptance sampling does not estimate the error rate directly, but can be used to determine whether to accept or reject batches of annotated instances.
For both we analysed the average sample sizes needed for two configurations of confidence and power.
We show that estimating the error rate using confidence intervals can require surprisingly large sample sizes, especially if more tight boundaries are wanted.
Acceptance sampling, especially sequential sampling with curtailment, can be a viable alternative as it overall requires far less inspection.
Finally, we urge researchers to apply statistical quality control during their annotation campaign, use large enough sample sizes and report their approaches alongside the recommended datasheet for datasets~\citep{gebruDatasheetsDatasets2021}. 
Ideally, the error rate of the final artifact is also estimated and reported.

\section{Limitations}

In this work, we described how to thoroughly estimate the error rate using confidence intervals and how to apply acceptance sampling for quality control during and after annotation campaigns.
While already working well, there are some limitations that need to be taken into account.
For our analysis, we used the closed form solutions to estimate the average sample numbers.
As these are theoretical results, it would be worthwhile to apply the presented methods during real annotation projects and study their feasibility there.
This is especially relevant for sequential sampling, as its good performance relies on true error rates that are either below the desired acceptance rate or above the rejection rate.
In cases where it is often in the indifference zone, it might degrade to single sampling.

It is also required that annotations are made independent and identically distributed.
This assumption needs to be validated in more depth, as annotations depend on the underlying datum to annotate which can be dependent, e.g., the text or image, or on the annotators.

Finally, one limiting factor and important reason why it is maybe only rarely deployed of quality control is the additional cost it comes with; it has to be weighed against annotating more with potentially having lower resulting data quality.

\clearpage

\bibliography{bibliography_clean,bibliography_raw}

\begin{thebibliography}{42}
\expandafter\ifx\csname natexlab\endcsname\relax\def\natexlab#1{#1}\fi

\bibitem[{Alex et~al.(2010)Alex, Grover, Shen, and Kabadjov}]{alexAgileCorpusAnnotation2010}
Bea Alex, Claire Grover, Rongzhou Shen, and Mijail Kabadjov. 2010.
\newblock Agile {{Corpus Annotation}} in {{Practice}}: {{An Overview}} of {{Manual}} and {{Automatic Annotation}} of {{CVs}}.
\newblock In \emph{Proceedings of the Fourth Linguistic Annotation Workshop}, pages 29--37, {Uppsala, Sweden}.

\bibitem[{Amidei et~al.(2019)Amidei, Piwek, and Willis}]{amideiAgreementOverratedPlea2019}
Jacopo Amidei, Paul Piwek, and Alistair Willis. 2019.
\newblock \href {https://doi.org/10.18653/v1/W19-8642} {Agreement is overrated: {{A}} plea for correlation to assess human evaluation reliability}.
\newblock In \emph{Proceedings of the 12th {{International Conference}} on {{Natural Language Generation}}}, pages 344--354, {Tokyo, Japan}.

\bibitem[{Aroian(1968)}]{aroianSequentialAnalysisDirect1968}
Leo~A. Aroian. 1968.
\newblock \href {https://doi.org/10.1080/00401706.1968.10490540} {Sequential {{Analysis}}, {{Direct Method}}}.
\newblock \emph{Technometrics}, 10(1):125--132.

\bibitem[{Artstein and Poesio(2008)}]{artsteinInterCoderAgreementComputational2008}
Ron Artstein and Massimo Poesio. 2008.
\newblock \href {https://doi.org/10.1162/coli.07-034-R2} {Inter-{{Coder Agreement}} for {{Computational Linguistics}}}.
\newblock \emph{Computational Linguistics}, 34(4):555--596.

\bibitem[{Banko and Brill(2001)}]{bankoScalingVeryVery2001}
Michele Banko and Eric Brill. 2001.
\newblock \href {https://doi.org/10.3115/1073012.1073017} {Scaling to {{Very Very Large Corpora}} for {{Natural Language Disambiguation}}}.
\newblock In \emph{Proceedings of the 39th {{Annual Meeting}} of the {{Association}} for {{Computational Linguistics}}}, pages 26--33, {Toulouse, France}.

\bibitem[{Barnes et~al.(2019)Barnes, {\O}vrelid, and Velldal}]{barnesSentimentAnalysisNot2019}
Jeremy Barnes, Lilja {\O}vrelid, and Erik Velldal. 2019.
\newblock \href {https://doi.org/10.18653/v1/W19-4802} {Sentiment {{Analysis Is Not Solved}}! {{Assessing}} and {{Probing Sentiment Classification}}}.
\newblock In \emph{Proceedings of the 2019 {{ACL Workshop BlackboxNLP}}: {{Analyzing}} and {{Interpreting Neural Networks}} for {{NLP}}}, pages 12--23, {Florence, Italy}.

\bibitem[{Bartroff et~al.(2022)Bartroff, Lorden, and Wang}]{bartroffOptimalFastConfidence2022}
Jay Bartroff, Gary Lorden, and Lijia Wang. 2022.
\newblock \href {https://doi.org/10.1080/00031305.2022.2128421} {Optimal and {{Fast Confidence Intervals}} for {{Hypergeometric Successes}}}.
\newblock \emph{The American Statistician}, 77(2):151--159.

\bibitem[{Brown et~al.(2001)Brown, Cai, and DasGupta}]{brownIntervalEstimationBinomial2001}
Lawrence~D. Brown, T.~Tony Cai, and Anirban DasGupta. 2001.
\newblock \href {https://doi.org/10.1214/ss/1009213286} {Interval {{Estimation}} for a {{Binomial Proportion}}}.
\newblock \emph{Statistical Science}, 16(2).

\bibitem[{Button et~al.(2013)Button, Ioannidis, Mokrysz, Nosek, Flint, Robinson, and Munaf{\`o}}]{buttonPowerFailureWhy2013}
Katherine~S. Button, John P.~A. Ioannidis, Claire Mokrysz, Brian~A. Nosek, Jonathan Flint, Emma S.~J. Robinson, and Marcus~R. Munaf{\`o}. 2013.
\newblock \href {https://doi.org/10.1038/nrn3475} {Power failure: Why small sample size undermines the reliability of neuroscience}.
\newblock \emph{Nature Reviews Neuroscience}, 14(5):365--376.

\bibitem[{Clopper and Pearson(1934)}]{clopperUseConfidenceFiducial1934}
C.~J. Clopper and E.~S. Pearson. 1934.
\newblock \href {https://doi.org/10.1093/biomet/26.4.404} {The use of confidence or fiducial limits illustrated in the case of the binomial}.
\newblock \emph{Biometrika}, 26(4):404--413.

\bibitem[{Dickinson(2005)}]{dickinsonErrorDetectionCorrection2005}
Markus Dickinson. 2005.
\newblock \emph{Error Detection and Correction in Annotated Corpora}.
\newblock Ph.D. thesis, The Ohio State University.

\bibitem[{Dodge(1943)}]{dodgeSamplingInspectionPlan1943}
H.~F. Dodge. 1943.
\newblock \href {https://doi.org/10.1214/aoms/1177731420} {A {{Sampling Inspection Plan}} for {{Continuous Production}}}.
\newblock \emph{The Annals of Mathematical Statistics}, 14(3):264--279.

\bibitem[{Duncan(1959)}]{duncanQualityControlIndustrial1959}
Acheson~J. Duncan. 1959.
\newblock \emph{Quality {{Control And Industrial Statistics}}}.
\newblock {Irwin Professional Publishing}.

\bibitem[{Gebru et~al.(2021)Gebru, Morgenstern, Vecchione, Vaughan, Wallach, Iii, and Crawford}]{gebruDatasheetsDatasets2021}
Timnit Gebru, Jamie Morgenstern, Briana Vecchione, Jennifer~Wortman Vaughan, Hanna Wallach, Hal~Daum{\'e} Iii, and Kate Crawford. 2021.
\newblock \href {https://doi.org/10.1145/3458723} {Datasheets for datasets}.
\newblock \emph{Communications of the ACM}, 64(12):86--92.

\bibitem[{Guenther(1969)}]{guentherUseBinomialHypergeometric1969}
William~C. Guenther. 1969.
\newblock \href {https://doi.org/10.1080/00224065.1969.11980358} {Use of the {{Binomial}}, {{Hypergeometric}} and {{Poisson Tables}} to {{Obtain Sampling Plans}}}.
\newblock \emph{Journal of Quality Technology}, 1(2):105--109.

\bibitem[{Hovy and Lavid(2010)}]{hovyScienceCorpusAnnotation2010}
Eduard Hovy and Julia Lavid. 2010.
\newblock Towards a `{{Science}}' of {{Corpus Annotation}}: {{A New Methodological Challenge}} for {{Corpus Linguistics}}.
\newblock \emph{International Journal of Translation Studies}, 22:13--36.

\bibitem[{Ide and Pustejovsky(2017)}]{ideHandbookLinguisticAnnotation2017}
Nancy Ide and James Pustejovsky, editors. 2017.
\newblock \href {https://doi.org/10.1007/978-94-024-0881-2} {\emph{Handbook of {{Linguistic Annotation}}}}.
\newblock {Springer Netherlands}, {Dordrecht}.

\bibitem[{{Institute of Medicine and National Research Council}(1985)}]{NAP372}
{Institute of Medicine and National Research Council}. 1985.
\newblock \href {https://doi.org/10.17226/372} {\emph{{An Evaluation of the Role of Microbiological Criteria for Foods and Food Ingredients}}}.
\newblock The National Academies Press, Washington, DC.

\bibitem[{Jain et~al.(2020)Jain, Patel, Nagalapatti, Gupta, Mehta, Guttula, Mujumdar, Afzal, Sharma~Mittal, and Munigala}]{jainOverviewImportanceData2020}
Abhinav Jain, Hima Patel, Lokesh Nagalapatti, Nitin Gupta, Sameep Mehta, Shanmukha Guttula, Shashank Mujumdar, Shazia Afzal, Ruhi Sharma~Mittal, and Vitobha Munigala. 2020.
\newblock \href {https://doi.org/10.1145/3394486.3406477} {Overview and {{Importance}} of {{Data Quality}} for {{Machine Learning Tasks}}}.
\newblock In \emph{Proceedings of the 26th {{ACM SIGKDD International Conference}} on {{Knowledge Discovery}} \& {{Data Mining}}}, pages 3561--3562, {Online}.

\bibitem[{Klie et~al.(2024)Klie, {de Castilho}, and Gurevych}]{klieAnalyzingDatasetAnnotation2024}
Jan-Christoph Klie, Richard~Eckart {de Castilho}, and Iryna Gurevych. 2024.
\newblock \href {http://arxiv.org/abs/2307.08153} {Analyzing {{Dataset Annotation Quality Management}} in the {{Wild}}}.
\newblock \emph{Computational Linguistics}.

\bibitem[{Klie et~al.(2023)Klie, Webber, and Gurevych}]{klieAnnotationErrorDetection2023}
Jan-Christoph Klie, Bonnie Webber, and Iryna Gurevych. 2023.
\newblock \href {https://doi.org/10.1162/coli_a_00464} {Annotation {{Error Detection}}: {{Analyzing}} the {{Past}} and {{Present}} for a {{More Coherent Future}}}.
\newblock \emph{Computational Linguistics}, 49(1):157--198.

\bibitem[{Krippendorff(2004)}]{krippendorffReliabilityContentAnalysis2004}
Klaus Krippendorff. 2004.
\newblock \href {https://doi.org/10.1111/j.1468-2958.2004.tb00738.x} {Reliability in {{Content Analysis}}.: {{Some Common Misconceptions}} and {{Recommendations}}}.
\newblock \emph{Human Communication Research}, 30(3):411--433.

\bibitem[{Luca et~al.(2020)Luca, Vandercappellen, and Claes}]{lucaWebbasedToolDesign2020}
Stijn Luca, Johan Vandercappellen, and Johan Claes. 2020.
\newblock \href {https://doi.org/10.1080/08982112.2019.1641207} {A web-based tool to design and analyze single- and double-stage acceptance sampling plans}.
\newblock \emph{Quality Engineering}, 32(1):58--74.

\bibitem[{Maas et~al.(2011)Maas, Daly, Pham, Huang, Ng, and Potts}]{maas-etal-2011-learning}
Andrew~L. Maas, Raymond~E. Daly, Peter~T. Pham, Dan Huang, Andrew~Y. Ng, and Christopher Potts. 2011.
\newblock \href {https://aclanthology.org/P11-1015} {Learning word vectors for sentiment analysis}.
\newblock In \emph{Proceedings of the 49th Annual Meeting of the Association for Computational Linguistics: Human Language Technologies}, pages 142--150, Portland, Oregon, USA.

\bibitem[{Meeker(1975)}]{meekerSequentialTestsHypergeometric1975}
William Meeker. 1975.
\newblock \emph{Sequential {{Tests}} of the {{Hypergeometric Distribution}}}.
\newblock Ph.D. thesis, Union College, {New York City, USA}.

\bibitem[{Monarch(2021)}]{monarchHumanintheLoopMachineLearning2021}
Robert Monarch. 2021.
\newblock \emph{Human-in-the-{{Loop Machine Learning}}: {{Active Learning}} and {{Annotation}} for {{Human-Centered AI}}}.
\newblock {Manning Publications}.

\bibitem[{Montgomery(2013)}]{montgomeryIntroductionStatisticalQuality2013}
Douglas~C. Montgomery. 2013.
\newblock \emph{Introduction to Statistical Quality Control}, 7th ed edition.
\newblock {Wiley}, {Hoboken, NJ}.

\bibitem[{Newcombe(1998)}]{newcombeTwosidedConfidenceIntervals1998}
R~G~G Newcombe. 1998.
\newblock Two-sided confidence intervals for the single proportion: Comparison of seven methods.
\newblock \emph{Statistics in medicine.}, 17(8).

\bibitem[{Northcutt et~al.(2021)Northcutt, Athalye, and Mueller}]{northcuttPervasiveLabelErrors2021}
Curtis~G. Northcutt, Anish Athalye, and Jonas Mueller. 2021.
\newblock \href {http://arxiv.org/abs/2103.14749} {Pervasive label errors in test sets destabilize machine learning benchmarks}.
\newblock In \emph{35th {{Conference}} on {{Neural Information Processing Systems Datasets}} and {{Benchmarks Track}}}, pages 1--13, {Online}.

\bibitem[{Passonneau and Carpenter(2014)}]{passonneauBenefitsModelAnnotation2014}
Rebecca~J. Passonneau and Bob Carpenter. 2014.
\newblock \href {https://doi.org/10.1162/tacl_a_00185} {The {{Benefits}} of a {{Model}} of {{Annotation}}}.
\newblock \emph{Transactions of the Association for Computational Linguistics}, 2:311--326.

\bibitem[{Pustejovsky and Stubbs(2013)}]{pustejovskyNaturalLanguageAnnotation2013}
J.~Pustejovsky and Amber Stubbs. 2013.
\newblock \emph{Natural Language Annotation for Machine Learning}.
\newblock {O'Reilly Media}, {Sebastopol, California, USA}.

\bibitem[{Reiss et~al.(2020)Reiss, Xu, Cutler, Muthuraman, and Eichenberger}]{reissIdentifyingIncorrectLabels2020}
Frederick Reiss, Hong Xu, Bryan Cutler, Karthik Muthuraman, and Zachary Eichenberger. 2020.
\newblock \href {https://doi.org/10.18653/v1/2020.conll-1.16} {Identifying {{Incorrect Labels}} in the {{CoNLL-2003 Corpus}}}.
\newblock In \emph{Proceedings of the 24th {{Conference}} on {{Computational Natural Language Learning}}}, pages 215--226, {Online}.

\bibitem[{Sambasivan et~al.(2021)Sambasivan, Kapania, Highfill, Akrong, Paritosh, and Aroyo}]{sambasivanEveryoneWantsModel2021}
Nithya Sambasivan, Shivani Kapania, Hannah Highfill, Diana Akrong, Praveen~Kumar Paritosh, and Lora~Mois Aroyo. 2021.
\newblock "{{Everyone}} wants to do the model work, not the data work": {{Data Cascades}} in {{High-Stakes AI}}.
\newblock In \emph{{{SIGCHI}}}, pages 1--21.

\bibitem[{Santner(1998)}]{santnerNoteTeachingBinomial1998}
Thomas~J. Santner. 1998.
\newblock \href {https://doi.org/10.5282/UBM/EPUB.1480} {A {{Note}} on {{Teaching Binomial Confidence Intervals}}}.
\newblock \emph{Teaching Statistic}, 20:20--23.

\bibitem[{Sun et~al.(2017)Sun, Shrivastava, Singh, and Gupta}]{sunRevisitingUnreasonableEffectiveness2017}
Chen Sun, Abhinav Shrivastava, Saurabh Singh, and Abhinav Gupta. 2017.
\newblock \href {https://doi.org/10.1109/ICCV.2017.97} {Revisiting {{Unreasonable Effectiveness}} of {{Data}} in {{Deep Learning Era}}}.
\newblock In \emph{2017 {{IEEE International Conference}} on {{Computer Vision}} ({{ICCV}})}, pages 843--852, {Venice, Italy}.

\bibitem[{Tantaratana and Thomas(1977)}]{tantaratanaTruncatedSequentialProbability1977}
Sawasd Tantaratana and John~B. Thomas. 1977.
\newblock \href {https://doi.org/10.1016/0020-0255(77)90050-0} {Truncated sequential probability ratio test}.
\newblock \emph{Information Sciences}, 13(3):283--300.

\bibitem[{Tjong Kim~Sang and De~Meulder(2003)}]{tjongkimsangIntroductionCoNLL2003Shared2003}
Erik~F. Tjong Kim~Sang and Fien De~Meulder. 2003.
\newblock Introduction to the {{CoNLL-2003}} shared task: {{Language-independent}} named entity recognition.
\newblock In \emph{Proceedings of the {{Seventh Conference}} on {{Natural Language Learning}} at {{HLT-NAACL}} 2003}, pages 142--147, {Edmonton, Alberta, Canada}.

\bibitem[{V{\u a}dineanu et~al.(2022)V{\u a}dineanu, Pelt, Dzyubachyk, and Batenburg}]{vadineanuAnalysisImpactAnnotation2022}
{\textcommabelow S}erban V{\u a}dineanu, Daniel Pelt, Oleh Dzyubachyk, and Joost Batenburg. 2022.
\newblock An {{Analysis}} of the {{Impact}} of {{Annotation Errors}} on the {{Accuracy}} of {{Deep Learning}} for {{Cell Segmentation}}.
\newblock In \emph{Proceedings of {{Machine Learning Research}}}, pages 1251--1267, {Honolulu, Hawaii, USA}.

\bibitem[{Vasudevan et~al.(2022)Vasudevan, Caine, {Gontijo-Lopes}, {Fridovich-Keil}, and Roelofs}]{vasudevanWhenDoesDough2022}
Vijay Vasudevan, Benjamin Caine, Raphael {Gontijo-Lopes}, Sara {Fridovich-Keil}, and Rebecca Roelofs. 2022.
\newblock When does dough become a bagel? {{Analyzing}} the remaining mistakes on {{ImageNet}}.
\newblock In \emph{Proceedings of the 36th {{Conference}} on {{Neural Information Processing Systems}}}, pages 1--15, {New Orleans, Louisiana, USA}.

\bibitem[{Voormann and Gut(2008)}]{voormannAgileCorpusCreation2008}
Holger Voormann and Ulrike Gut. 2008.
\newblock \href {https://doi.org/10.1515/CLLT.2008.010} {Agile corpus creation}.
\newblock \emph{Corpus Linguistics and Linguistic Theory}, 4(2).

\bibitem[{Wald(1947)}]{waldSequentialAnalysis1947}
Abraham Wald. 1947.
\newblock \emph{Sequential {{Analysis}}}.
\newblock {John Wiley and Sons}, {New York}.

\bibitem[{Wang(2015)}]{wangExactOptimalConfidence2015}
Weizhen Wang. 2015.
\newblock \href {https://doi.org/10.1080/01621459.2014.966191} {Exact {{Optimal Confidence Intervals}} for {{Hypergeometric Parameters}}}.
\newblock \emph{Journal of the American Statistical Association}, 110(512):1491--1499.

\end{thebibliography}

\clearpage

\appendix
\onecolumn

\section{Acceptance Sampling Simulations using Existing NLP Datasets}
\label{sec:app_real_app}

This section contains additional information extending the results from the experiments in \cref{sec:app_real}.

\begin{table}[h]
\begin{tabular}{@{}lllr@{}}
\toprule
\textbf{Dataset}                      & \textbf{QA Config}                & \textbf{Plan}             & \textbf{Sample Size} \\ \midrule
\multirow{14}{*}{CoNLL 2003} & \multirow{7}{*}{strict}  & Single Sampling           & 585                             \\
                             &                          & Double Sampling Full      & 720                             \\
                             &                          & Double Sampling Curtailed & 361.296                         \\
                             &                          & Sequential Sampling Full          & 92.36                           \\
                             &                          & Sequential Sampling Curtailed     & 92.36                           \\
                             &                          & Confidence Interval $p_a$          & 803                             \\
                             &                          & Confidence Interval $p_r$         & 1389                            \\ \cmidrule(l){2-4} 
                             & \multirow{7}{*}{relaxed} & Single Sampling           & 278                             \\ 
                             &                          & Double Sampling Full      & 314                             \\
                             &                          & Double Sampling Curtailed & 184.508                         \\
                             &                          & Sequential Sampling Full          & 91.833                          \\
                             &                          & Sequential Sampling Curtailed     & 98.299                          \\
                             &                          & Confidence Interval $p_a$          & 288                             \\
                             &                          & Confidence Interval $p_r$         & 443                             \\ \midrule
\multirow{14}{*}{IMDB}       & \multirow{7}{*}{strict}  & Single Sampling           & 682                             \\
                             &                          & Double Sampling Full      & 758                             \\
                             &                          & Double Sampling Curtailed & 682.675                         \\
                             &                          & Sequential Sampling Full          & 627.084                         \\
                             &                          & Sequential Sampling Curtailed     & 461.713                         \\
                             &                          & Confidence Interval $p_a$          & 992                             \\
                             &                          & Confidence Interval $p_r$         & 1993                            \\ \cmidrule(l){2-4}
                             & \multirow{7}{*}{relaxed} & Single Sampling           & 305                             \\
                             &                          & Double Sampling Full      & 350                             \\
                             &                          & Double Sampling Curtailed & 347.572                         \\
                             &                          & Sequential Sampling Full          & 620.491                         \\
                             &                          & Sequential Sampling Curtailed     & 154.949                         \\
                             &                          & Confidence Interval $p_a$          & 260                             \\
                             &                          & Confidence Interval $p_r$         & 501                             \\ \bottomrule
\end{tabular}
\caption{(Average) sample sizes when applying acceptance sampling and using confidence intervals on ConLL 2003 and IMDB over 1000 repetitions. Overall, sequential sampling with curtailment has the lowest samples inspected.}
\label{tab:real-world}
\end{table}

\clearpage

\section{Validation}

Acceptance sampling for a hypergeometric distribution is quite intricate and difficult to implement.
We validate our implementations for the different acceptance sampling approaches in the following and show that our implementation is correct.

\subsection{Comparison of Operating Characteristic curves}

The Operating Characteristic curve (OC) for an acceptance plan shows the ability to discriminate good and bad datasets over the range of possible error rates.
When using the same parameters for a binomial and hypergeometric plan and simulating sampling with replacement for the binomial and with replacement for the hypergeometric case, the OC curves of both should be close to each other.
The binomial solution has available implementations and closed forms that can be assumed to be correct, we re-implemented these and compared our implementation to existing implementations and tabulated values.
This binomial case is then used to compare OC curves to acceptance sampling with the hypergeometric distribution.
OC curves for single, double, and sequential sampling across different dataset sizes can be seen in \cref{fig:ocr_valid}.
Curves for binomial and hypergeometric as the underlying distribution follow each other very close, validating our implementations.

\begin{figure}[H]
    \centering
    \captionsetup[subfigure]{justification=centering}
    \begin{subfigure}{\textwidth}
        \centering
        \includegraphics[width=\textwidth]{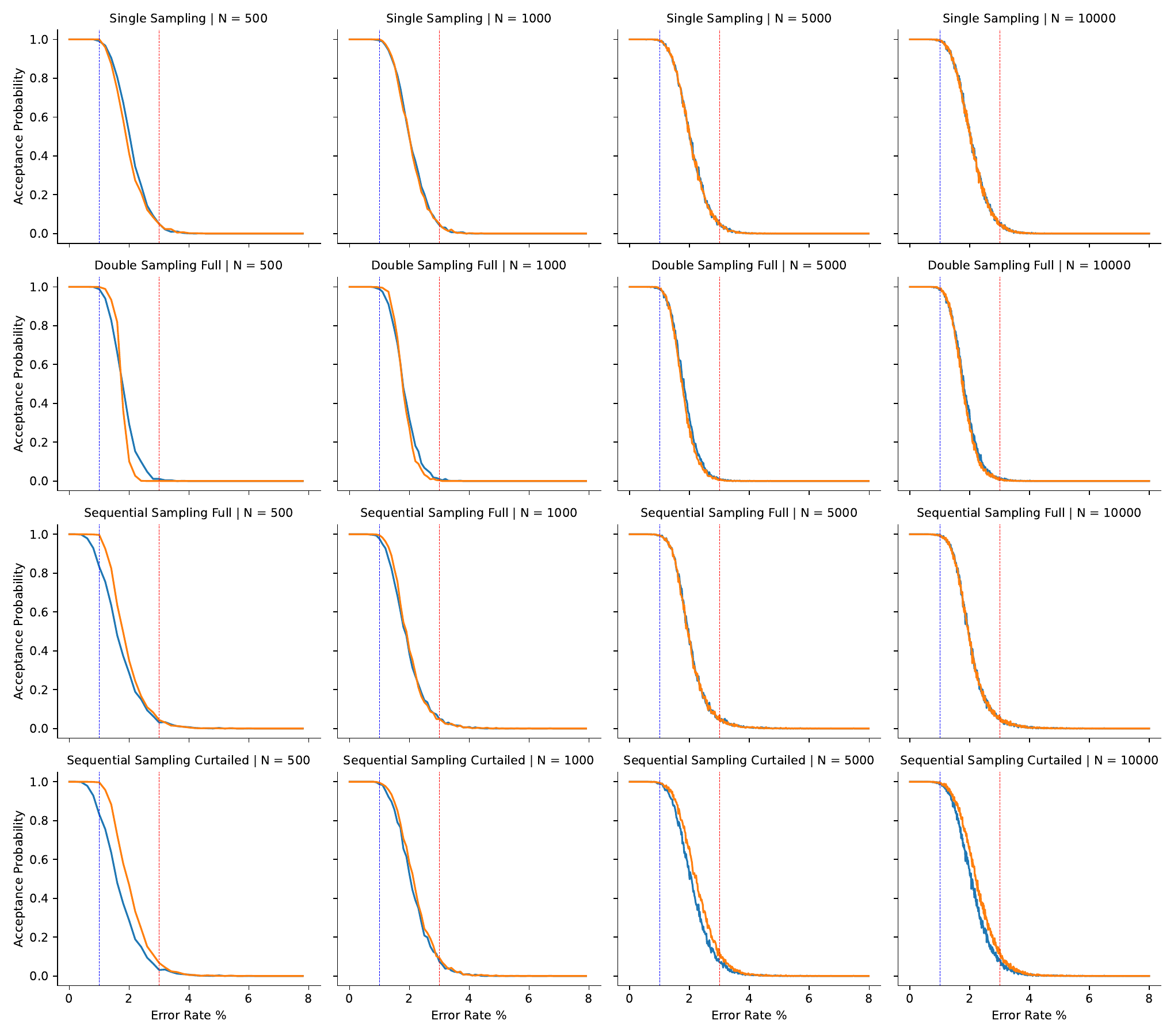}
        \caption{Strict}
     \end{subfigure}
\end{figure}

\begin{figure}[H]
    \ContinuedFloat
    \begin{subfigure}{\textwidth}
        \centering
        \includegraphics[width=\textwidth]{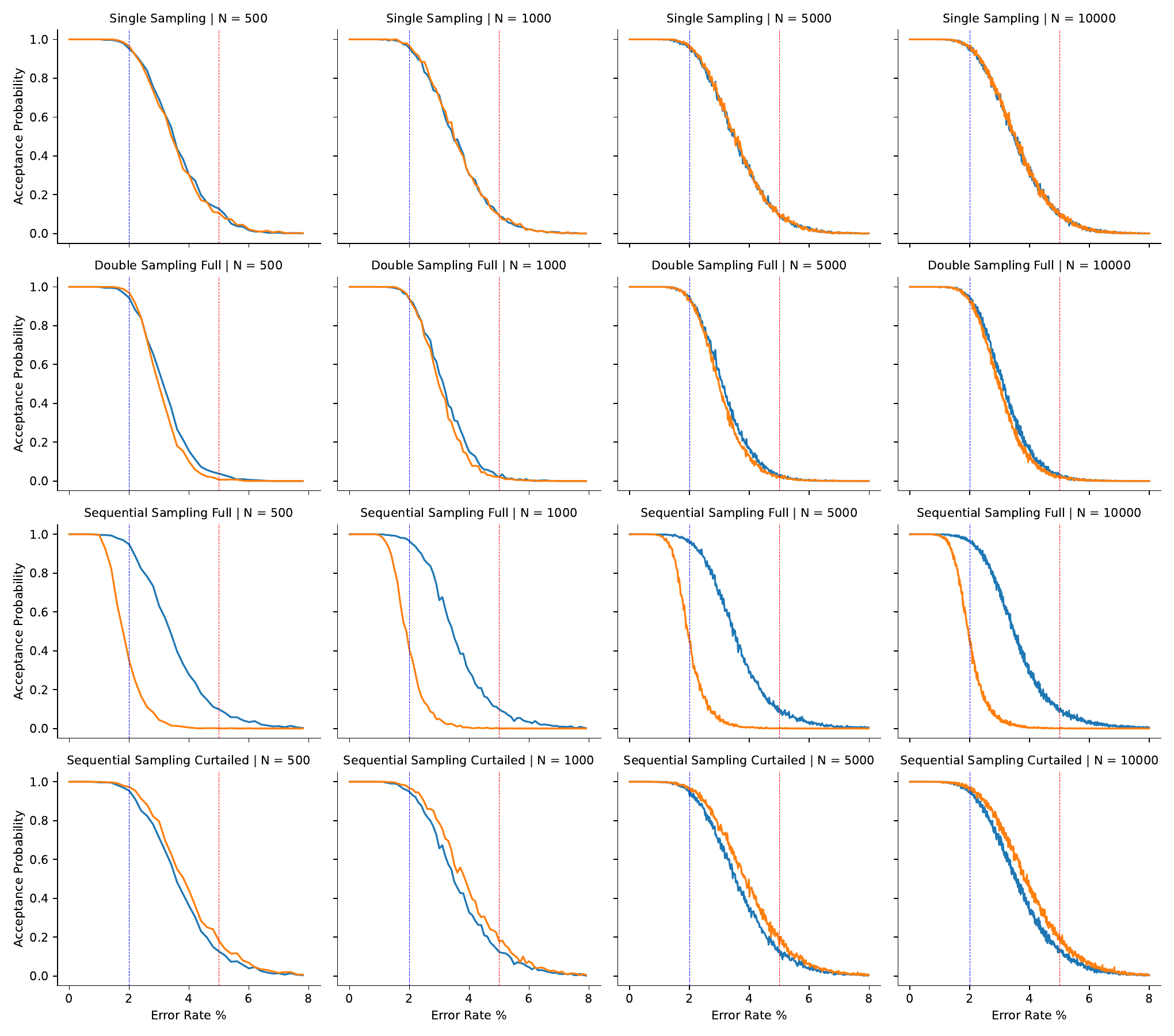}
        \caption{Relaxed}
    \end{subfigure}
    \caption{Operating Characteristic curves plotting error rate against probability of accepting a lot. We simulate acceptance sampling with replacement (binomial \tikzcircle[fill=blue]{3pt}) and without replacement (hypergeometric \tikzcircle[fill=orange]{3pt}) over 1000 repetitions and count how often for a probability, plans told to accept or reject.
    Horizontal lines indicate $p_a$ \tikzcircle[fill=blue]{3pt} and $p_r$ \tikzcircle[fill=red]{3pt}.
    Note that there is no real stopping criterion for sequential sampling for binomial, which is why we simulate stopping either if the number of errors steps outside the boundaries or after 10 times the dataset size.}
    \label{fig:ocr_valid}
\end{figure}

\clearpage

\subsection{Simulation}

\begin{figure}[H]
    \centering
    \captionsetup[subfigure]{justification=centering, belowskip=.5em}
    \begin{subfigure}{0.45\textwidth}
        \centering
        \includegraphics[width=\textwidth]{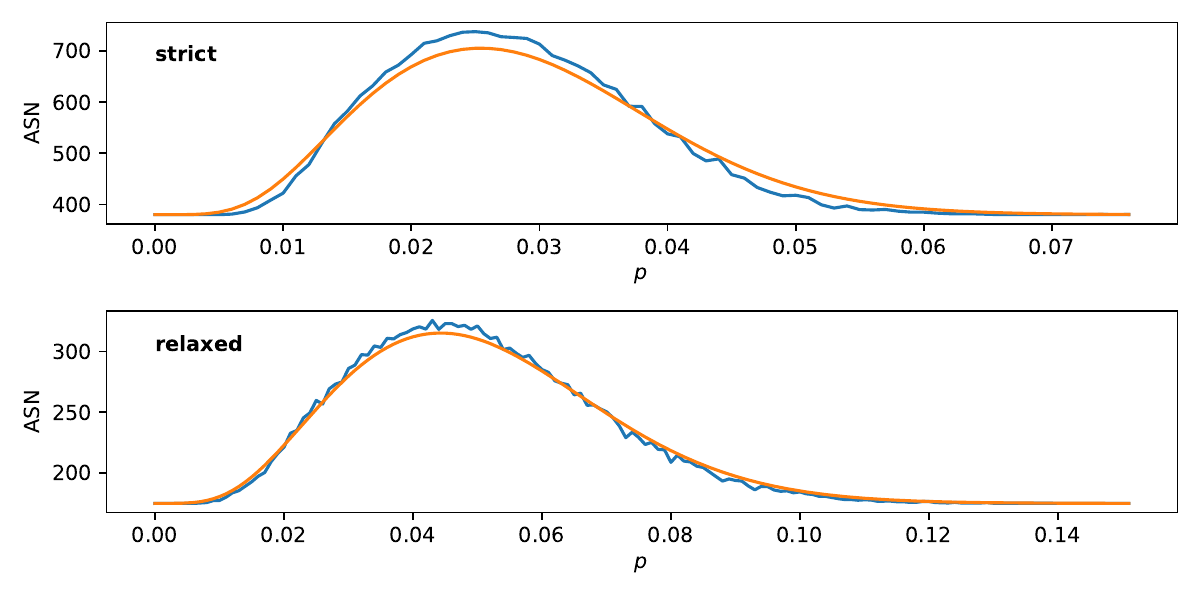}
        \caption{Double Sampling Full Binomial}
        \label{fig:sim_double_binomial_full}
     \end{subfigure}
    \begin{subfigure}{0.45\textwidth}
        \centering
        \includegraphics[width=\textwidth]{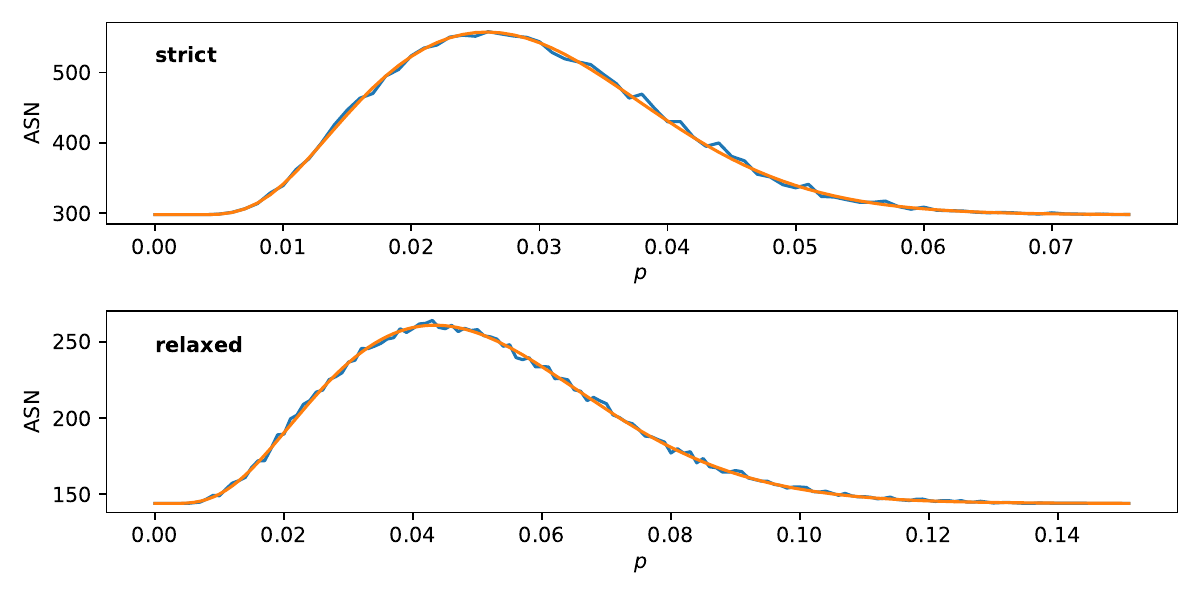}
        \caption{Double Sampling Full Hypergeometric}
        \label{fig:sim_double_hypergeometric_full}
     \end{subfigure}

    \centering
    \begin{subfigure}{0.45\textwidth}
        \centering
        \includegraphics[width=\textwidth]{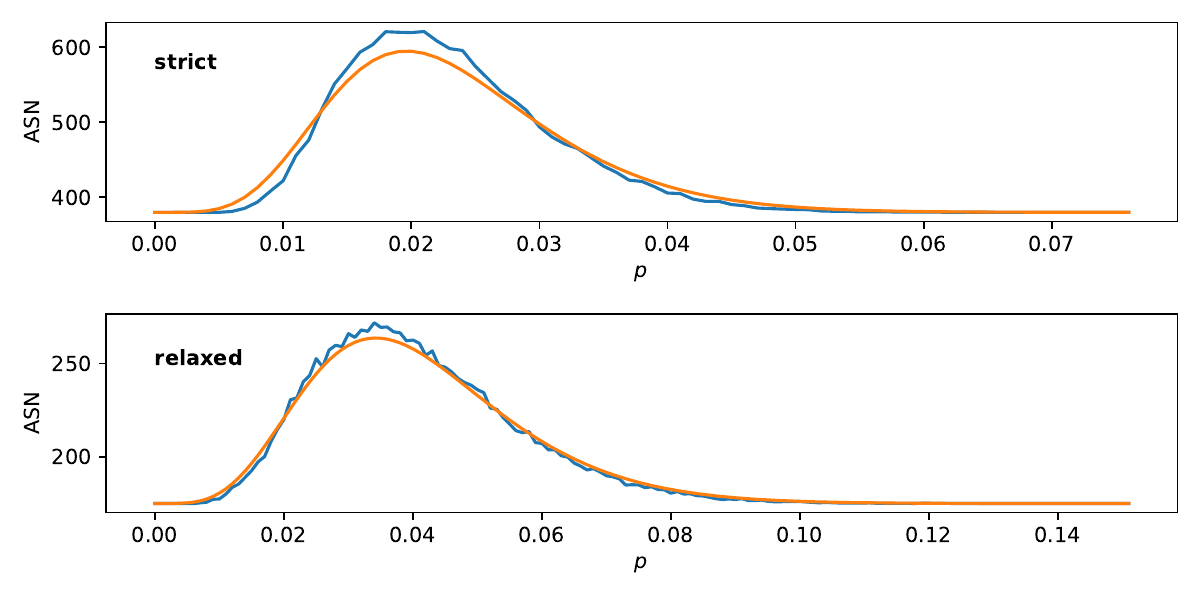}
        \caption{Double Sampling Curtailed Binomial}
        \label{fig:sim_double_binomial_curtailed}
    \end{subfigure}
    \centering
    \begin{subfigure}{0.45\textwidth}
        \centering
        \includegraphics[width=\textwidth]{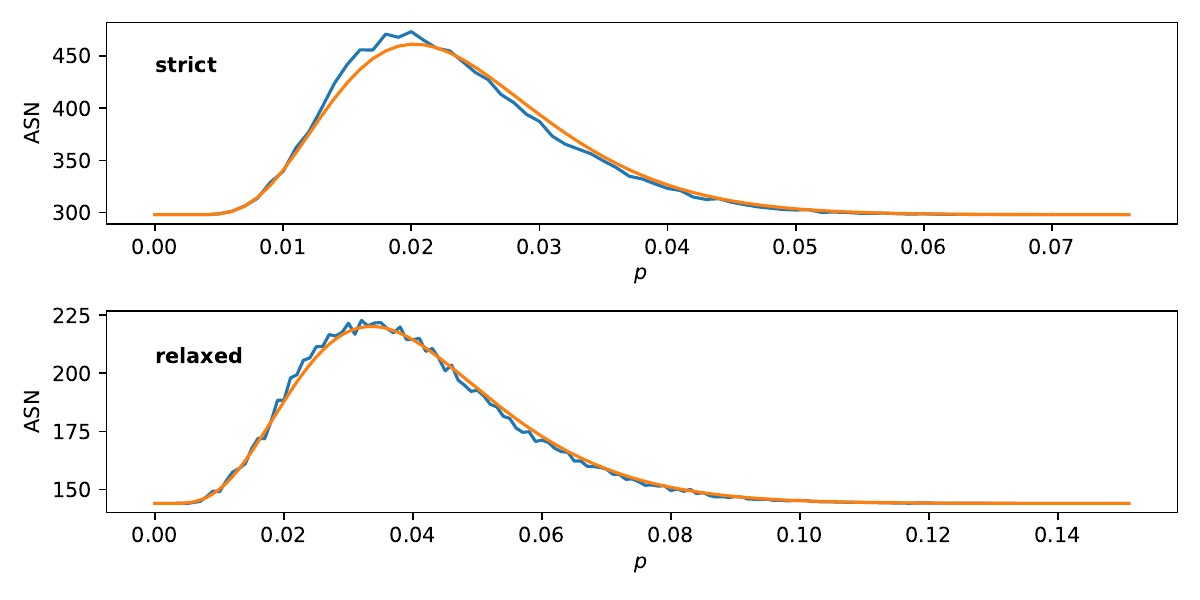}
        \caption{Double Sampling Curtailed Hypergeometric}
        \label{fig:sim_double_hypergeometric_curtailed}
    \end{subfigure}

    \centering
    \begin{subfigure}{0.45\textwidth}
        \centering
        \includegraphics[width=\textwidth]{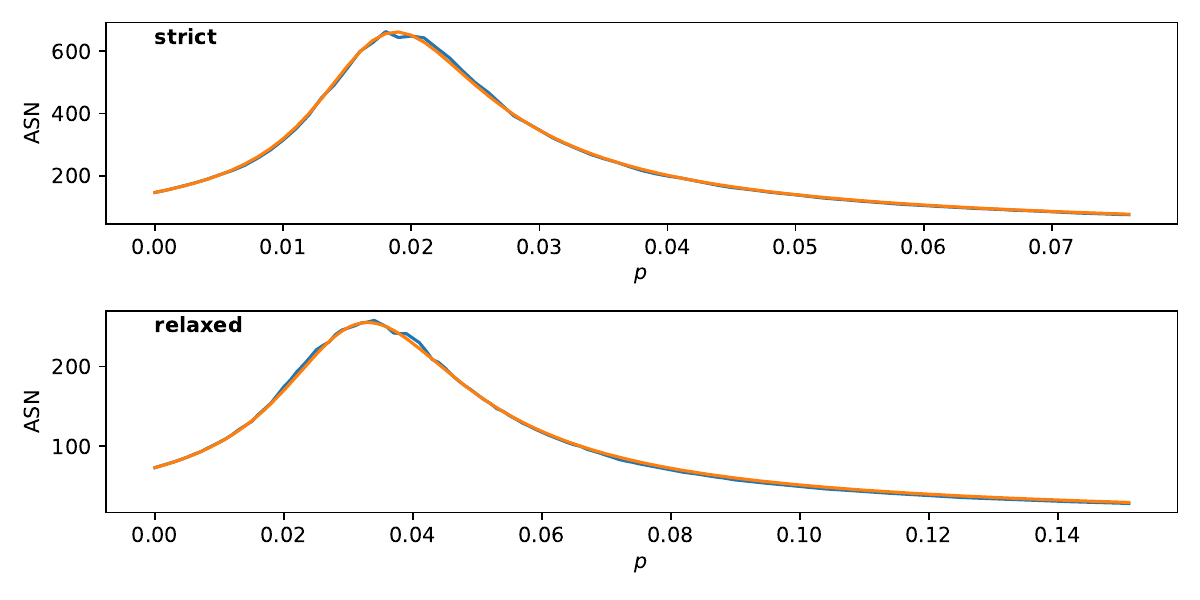}
        \caption{Sequential Sampling Full Hypergeometric}
        \label{fig:sim_sequential_binomial_full}
     \end{subfigure}
    \centering
    \begin{subfigure}{0.45\textwidth}
        \centering
        \includegraphics[width=\textwidth]{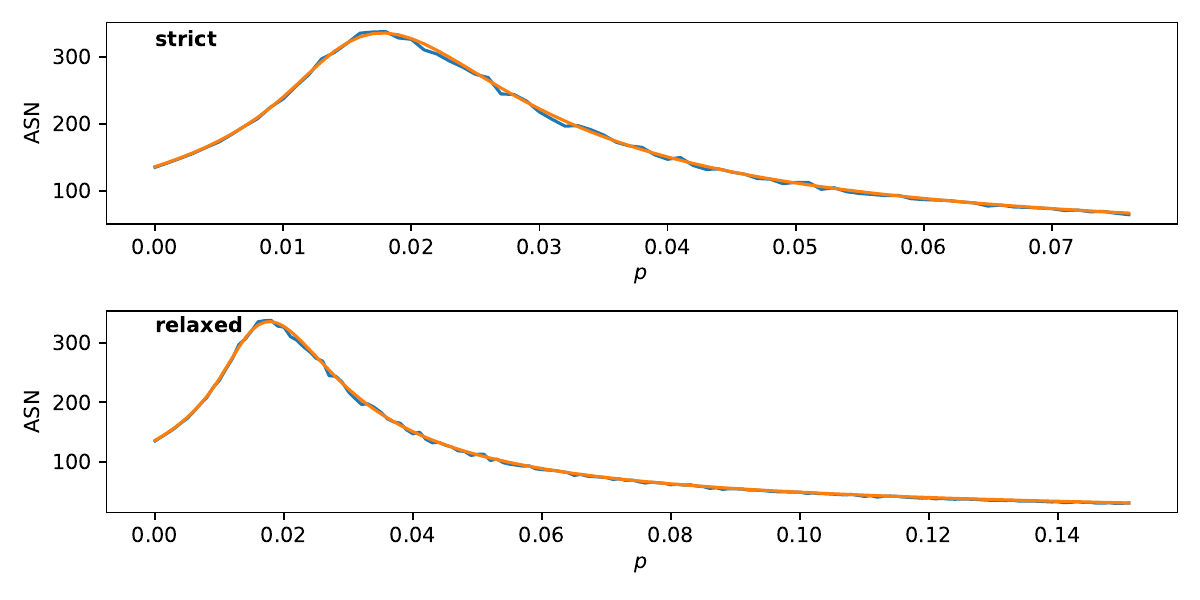}
        \caption{Sequential Sampling Full Hypergeometric}
        \label{fig:sim_sequential_hypergeometric_full}
     \end{subfigure}     
     \phantomcaption
     \label{fig:sim_vs_analytical1}

    \centering
    \begin{subfigure}{0.45\textwidth}
        \centering
        \includegraphics[width=\textwidth]{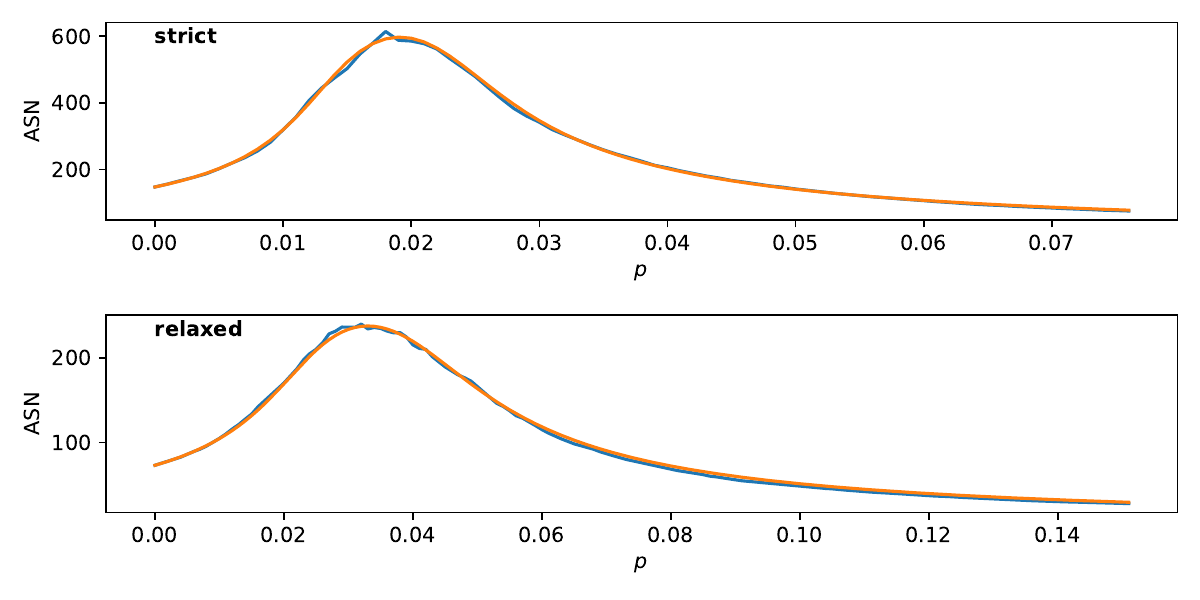}
        \caption{Sequential Sampling Curtailed Binomial}
        \label{fig:sim_sequential_binomial_curtailed}
     \end{subfigure}
     \centering
    \begin{subfigure}{0.45\textwidth}
        \centering
        \includegraphics[width=\textwidth]{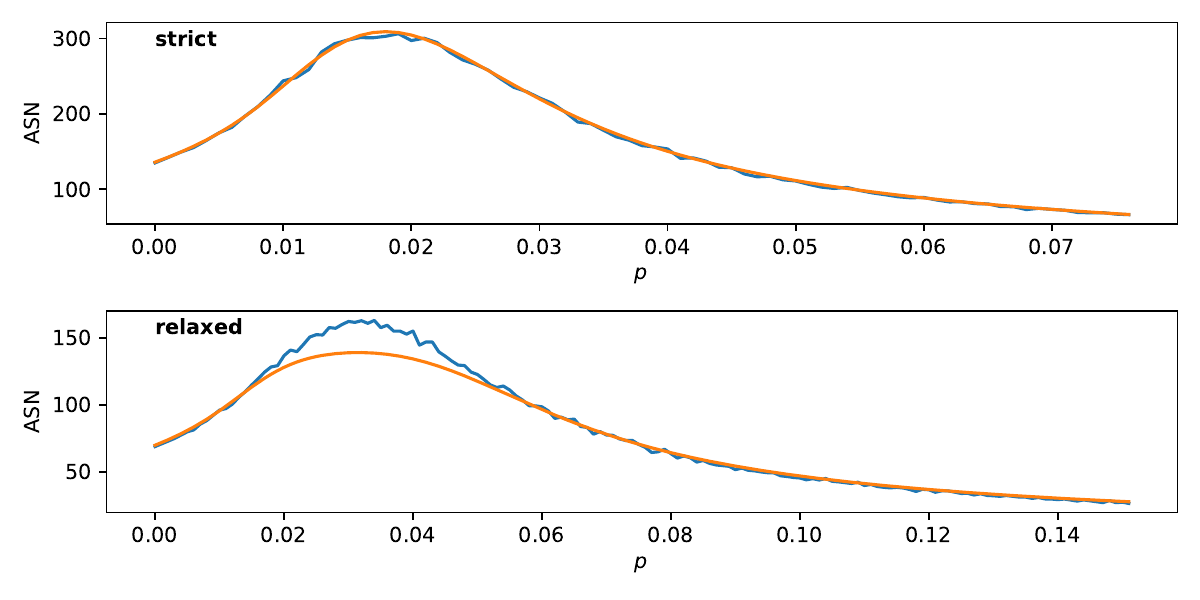}
        \caption{Sequential Sampling Curtailed Hypergeometric}
        \label{fig:sim_sequential_hypergeometric_curtailed}
     \end{subfigure}     
    \caption{Analytical \tikzcircle[fill=orange]{3pt} and \tikzcircle[fill=blue]{3pt}  simulated results for the average sampling numbers (ASN) across approaches and error rates $p$.  
    We simulate 1000 samples per possible error count in $0\ldots1000$. 
    It can be seen that the simulated sample size numbers are almost always very close to the analytically obtained ones, validating that our implementations are correct.  
    In the case of the curtailed sequential sampling, the average sample numbers can be slightly overestimated in the indifference zone, which we attribute to the approximate curtailment used. }
    \label{fig:sim_vs_analytical}
\end{figure}

We repeatedly simulate manual inspection as sampling without replacement, compute the \acf{asn} and compare the results to the analytical solutions.
The results are depicted in \cref{fig:sim_vs_analytical1}.
It can be seen that both are almost identical, showing the correctness of our implementation.
We attribute minor differences for sequential sampling to the approximated curtailment in the analytical computation.

\clearpage

\section{Using the Hypergeometric vs. the Binomial Distribution}
\label{app:binomial}

Manual inspection as sampling without replacement is best described by the hypergeometric distribution.
It is often approximated by the binomial distribution, as for the latter, more implementations and research is available.
In \cref{fig:asn_p_binom}, we compare both across different dataset sizes and show that approximation can lead to inaccuracies.
Therefore, care needs to be taken to only approximate if the dataset size is at least 10 times larger than the sample size.
For the configurations used in this work, this turns out to be dataset sizes above 5000 instances.

\vspace{4em}

\begin{figure}[h]
    \centering
    \captionsetup[subfigure]{justification=centering, aboveskip=5pt,belowskip=15pt}

    \begin{subfigure}[t]{0.99\textwidth}
        \centering
        \includegraphics[width=\textwidth]{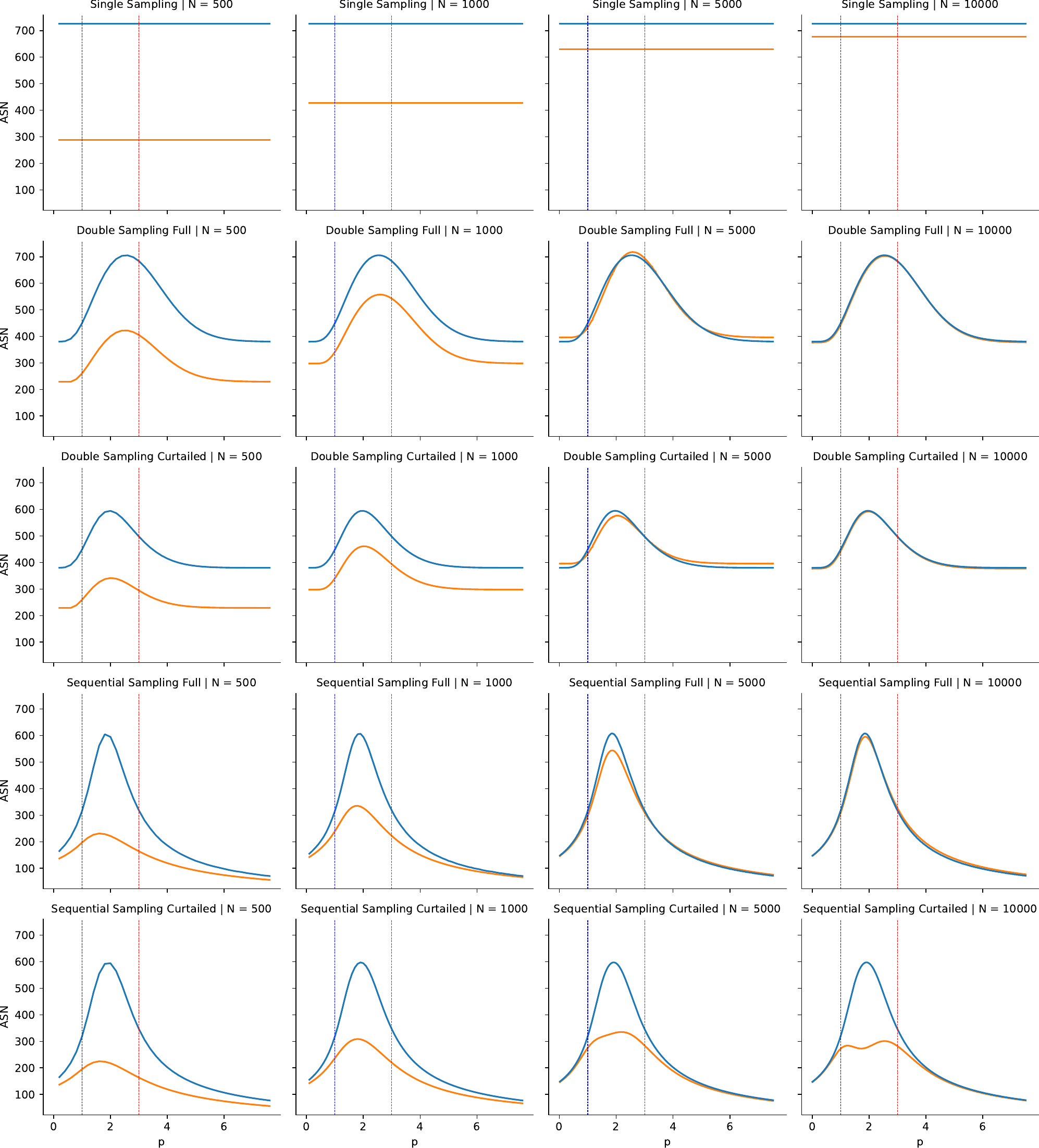}
        \caption{Strict}
     \end{subfigure}
     \hfill
\end{figure}
\begin{figure}
    \ContinuedFloat
    \begin{subfigure}[t]{0.99\textwidth}
        \centering
        \includegraphics[width=\textwidth]{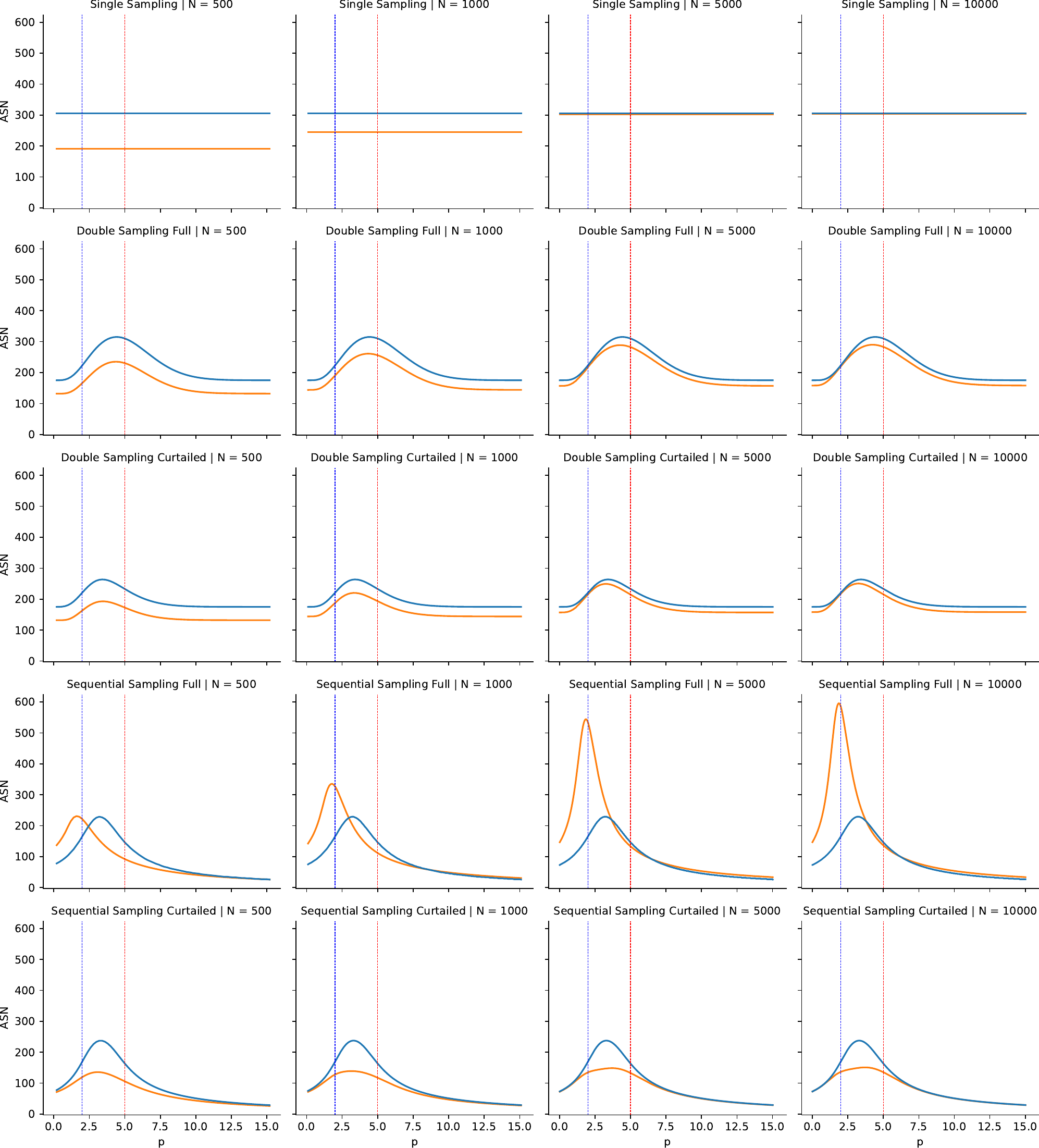}
        \caption{Relaxed}
     \end{subfigure}
    \caption{Comparing the analytical average sample numbers (ASN) when using the binomial distribution (sampling with replacement, \tikzcircle[fill=blue]{3pt}) and hypergeometric distribution (sampling with replacement, \tikzcircle[fill=orange]{3pt}) to compare the average sample numbers required for a strict and relaxed configuration across different dataset sizes $N$. 
    The hypergeometric model best describes annotation inspection, as no instance is inspected twice; the binomial is an approximation. It can be seen that there are stark differences between the suggested average sample numbers, especially for dataset sizes below $5000$ and for sequential sampling.
    Horizontal lines indicate $p_a$ \tikzcircle[fill=blue]{3pt} and $p_r$ \tikzcircle[fill=red]{3pt}.}
    \label{fig:asn_p_binom}
\end{figure}

\end{document}